\documentclass[sn-mathphys-num]{sn-jnl}

\usepackage{graphicx}%
\usepackage{multirow}%
\usepackage{amsmath,amssymb,amsfonts}%
\usepackage{amsthm}%
\usepackage{mathrsfs}%
\usepackage[title]{appendix}%
\usepackage{xcolor}%
\usepackage{textcomp}%
\usepackage{manyfoot}%
\usepackage{booktabs}%
\usepackage{algorithm}%
\usepackage{algorithmicx}%
\usepackage{algpseudocode}%
\usepackage{listings}%
\usepackage{adjustbox}
\usepackage{array}
\usepackage{tabularx}
\usepackage{makecell}
\usepackage{diagbox}
\usepackage{multirow}
\usepackage{textcomp}
\usepackage{xcolor}

\usepackage{adjustbox}
\usepackage{tikz}
\usetikzlibrary{bayesnet, positioning, arrows.meta, shapes}
\usetikzlibrary{
    shapes.geometric, 
    shapes.misc,      
    positioning,      
    arrows.meta,      
    calc              
}
\theoremstyle{thmstyleone}%
\theoremstyle{thmstyletwo}%

\theoremstyle{thmstylethree}%

\begin{document}



\title{Generative AI in Depth: A Survey of Recent Advances, Model Variants, and Real-World Applications}




\author{Shamim Yazdani$^1$, Akansha Singh$^2$, Nripsuta Saxena$^3$, Zichong Wang$^1$, Avash Palikhe$^1$, Deng Pan$^1$, Umapada Pal$^4$, Jie Yang$^5$}

\affil[1]{\orgname{Florida International University}, \city{Miami}, \state{Florida}, \country{USA}}
\affil[2]{\orgname{Manipal Institute of Technology}, \state{Karnataka}, \country{India}}
\affil[3]{\orgname{University of Southern California}, \orgaddress{\street{California}, \country{USA}}}
\affil[4]{\orgname{Indian Statistical Institute}, \orgaddress{\street{Kolkata}, \country{India}}}
\affil[5]{\orgname{University of Wollongong}, \orgaddress{\street{Wollongong}, \country{Australia}}}

 \author*[1]{\fnm{Wenbin} \sur{Zhang}}\email{wenbin.zhang@fiu.edu}



\abstract{

In recent years, deep learning based generative models, particularly Generative Adversarial Networks (GANs), Variational Autoencoders (VAEs), and Diffusion Models (DMs), have been instrumental in in generating diverse, high-quality content across various domains, such as image and video synthesis. This capability has led to widespread adoption of these models and has captured strong public interest. As they continue to advance at a rapid pace, the growing volume of research, expanding application areas, and unresolved technical challenges make it increasingly difficult to stay current. To address this need, this survey introduces a comprehensive taxonomy that organizes the literature and provides a cohesive framework for understanding the development of GANs, VAEs, and DMs, including their many variants and combined approaches. We highlight key innovations that have improved the quality, diversity, and controllability of generated outputs, reflecting the expanding potential of generative artificial intelligence. In addition to summarizing technical progress, we examine rising ethical concerns, including the risks of misuse and the broader societal impact of synthetic media. Finally, we outline persistent challenges and propose future research directions, offering a structured and forward looking perspective for researchers in this fast evolving field.
}


\maketitle

\section{Introduction}

{T}{he} field of generative artificial intelligence (AI) has made significant strides within the swiftly evolving landscape of AI~\cite{gozalo2023survey,liu2021generative,jin2023rethinking,amon2024uncertain}. The capability of generative AI models to produce realistic images and videos has clear and noteworthy practical implications across a multitude of domains, such as art~\cite{elgammal2017can, li2017universal}, entertainment \cite{rht_kum_sin-25a}, gaming~\cite{liu2021deep}, design~\cite{gatys2016image}, medicine~\cite{pinaya2023generative, jing2023eigenfold, shokrollahi2023comprehensive, ktena2023generative}, and data augmentation~\cite{perez2017effectiveness} for machine learning tasks. Leading the revolution in generative AI are the continuous surge in computing power and the emergence of innovative deep learning (DL) architectures such as Generative Adversarial Networks (GANs)~\cite{goodfellow2014generative}, Variational Autoencoders (VAEs)~\cite{kingma2013auto, rezende2014stochastic}, and Diffusion Models (DMs)~\cite{karras2023analyzing, wu2023fast, yang2023diffusion, croitoru2023diffusion, de2021diffusion}, 
which excel at capturing intricate data patterns and producing more realistic and varied images, surpassing traditional generative methods~\cite{bengesi2023advancements}.




Although just a decade old, these architectures have transformed the realm of generative AI in various domains. 
Introduced in 2014, GANs transformed the training of generative models with a new revolutionary framework consisting of two essential components: a generator and a discriminator. Locked in an adversarial competition~\cite{goodfellow2014generative}, the generator produces new samples, while the discriminator aims to distinguish between fake and real images.
By incorporating an adversarial competition between the generator and discriminator, GANs can produce remarkably realistic and diverse image and video samples~\cite{ali2021gans}. 
GANs have found applications in various areas within the realm of visual media generation, including but not limited to image-to-image translation~\cite{isola2017image}, object detection~\cite{zhang2020multi}, artistic style transferring~\cite{xu2021drb}, and even gesture generation in robots~\cite{nishimura2020human}. 
On the other hand, VAEs leverage the capabilities of both autoencoders and probabilistic modeling to acquire insightful data representations within a latent space. In contrast to developing an encoder that generates a solitary value for each latent attribute, the VAE's encoder is crafted to articulate a probability distribution for each latent attribute~\cite{kingma2013auto}. This design 
facilitates seamless interpolation, meaningful sampling, and the generation of high-quality images. VAEs find application 
in image and video generation~\cite{gregor2015draw,yan2021videogpt}, classification~\cite{gregor2015draw}, and environmental simulation~\cite{cheng2023generative}.
The most recent among the three architectures, DMs have revolutionized data-driven image and video synthesis, establishing themselves as leaders in managing large datasets~\cite{karras2023analyzing}, evidenced by their remarkable proficiency in capturing intricate data distributions~\cite{wu2023fast}. DMs consist of two key phases: a forward diffusion stage and a reverse diffusion stage. 
The input data undergoes gradual perturbation with Gaussian noise across multiple steps in the forward phase, and the model systematically reconstructs the original input by reversing the diffusion process in the reverse stage.
Artistic painting and text-guided image editing~\cite{croitoru2023diffusion}, MRI reconstruction~\cite{gungor2023adaptive}, self-driving cars~\cite{hu2023gaia}, and video production from text~\cite{croitoru2023diffusion} are among the successful applications of DMs in the domain of image and video generation.\par
Despite their transformative impact on generative AI, models such as GANs, VAEs, and DMs continue to face notable limitations. For instance, GANs demand meticulous stabilization during training \cite{goodfellow2014generative,arjovsky2017wasserstein}, as the adversarial nature of the generator and discriminator optimized via alternating or simultaneous gradient descent can easily lead to instability, mode collapse, or non-convergence. VAEs, while offering a principled probabilistic framework, often produce blurry outputs when modeling complex data distributions, due to their reliance on pixel-wise reconstruction losses \cite{dos_bro-16a, larsen2016autoencoding}. In contrast, Diffusion Models achieve impressive sample quality but at the cost of computational inefficiency, requiring hundreds to thousands of denoising steps to generate a single sample. To overcome these challenges, a wide range of strategies have been developed to improve training stability, enhance sample quality and diversity, and reduce inference time \cite{arjovsky2017wasserstein, radford2015unsupervised, cheng2023generative, isola2017image, zhao2017infovae, cao2022exploring, yang2023diffusion}. The resulting surge in research has significantly expanded the field, making it increasingly difficult to stay current with ongoing advancements and unresolved questions. To this end, this paper presents a comprehensive survey systematically covering the advancements in generative AI architectures as well as improvements and limitations. Specifically, \textit{this work, to the best of our knowledge, presents the first thorough study of the foundational breakthroughs, continued advancements, and current shortcomings in generative AI explicitly focusing on GANs, VAEs and DMs, for researchers and practitioners to have a thorough understanding of the current landscape.} The primary contributions of this paper include:

\begin{enumerate}
    \item We introduce a novel taxonomy of generative models, spanning GANs, VAEs, hybrid GAN-VAE architectures, and Diffusion Models (DMs) that categorizes them based on key design principles, offering a structured lens through which to understand the evolution of the field.
    
    \item Building on this taxonomy, we provide a systematic and comprehensive review of these models, addressing core challenges such as high quality and diverse image and video generation, as well as stable training dynamics.
    
    \item Our survey offers an in-depth comparison of the methodologies, strengths, and limitations associated with GANs, VAEs, GAN-VAE hybrids, and DMs, along with their many proposed variants developed for visual content synthesis.
    
    \item In contrast to existing surveys, our work uniquely integrates technical advancements with a systematic classification, while also exploring real world applications across domains enabled by model improvements.
    
    \item We examine the ethical implications and societal risks posed by generative AI, and identify open challenges and future research directions to support responsible innovation in this rapidly evolving area.    
\end{enumerate}

The remainder of this paper is organized as follows. The taxonomy is presented in Section \ref{taxonomy}. Section \ref{GANs} explores GANs, followed by an in-depth discussion of VAEs in Section \ref{VAEs}. Diffusion models are detailed in Section \ref{Diffusion}, and GAN-VAE hybrids are introduced in Section \ref{Hybrids}. 
The myriad applications of these approaches are covered in Section \ref{apps}, and ethical considerations specific to generative AI are discussed in Section \ref{ethics}. Section \ref{future} outlines challenges and directions for future work, and we conclude with Section \ref{conclusion}.
\begin{table*}[htbp]
\scriptsize 
\renewcommand{\arraystretch}{0.9} 
\setlength{\tabcolsep}{3pt} 
\centering
\resizebox{\textwidth}{!}{%
\begin{tabular}{>{\centering\arraybackslash}p{2cm} >{\centering\arraybackslash}p{2cm} >{\centering\arraybackslash}p{2.2cm} >{\centering\arraybackslash}p{3.2cm} >{\centering\arraybackslash}p{5cm}}
\hline
\textbf{Paper} & \textbf{Type} & \textbf{Method} & \textbf{Applications} & \textbf{Datasets} \\
\hline
    Isola et al.~\cite{isola2017image} & GAN & Conditional GAN & Image-to-image translation & Cityscapes dataset~\cite{cordts2016cityscapes}\\

     &  &  & semantic segmentation & CMP Facades~\cite{tylevcek2013spatial}, binary edges generated, using the HED edge detector~\cite{xie2015holistically}, data scraped from Google Maps, ~\cite{russakovsky2015imagenet}, ~\cite{yu2014fine}, ~\cite{zhu2016generative}, ~\cite{eitz2012humans}, ~\cite{laffont2014transient}\\

    \hline
    Zhang et al.~\cite{zhang2020multi} & GAN & MTGAN & object detection & COCO~\cite{lin2014microsoft}and WIDER FACE datasets~\cite{yang2016wider} \\
    \hline
    Croitoru et al.~\cite{croitoru2023diffusion} & Diffusion Models & DDPM & artistic paintings and text-guided image editing & CelebA-HQ, ImageNet~\cite{lugmayr2022repaint}, PaintByWord~\cite{avrahami2022blended}\\
    & &ADM & artistic paintings and text-guided image editing & MS-COCO~\cite{nichol2021glide} \\ 
     &  &FDM  & video production from text & GQN-Mazes, MineRL Navigate,CARLA Town01~\cite{harvey2022flexible}\\
     &  & DDPM & video production from text & 101 Human Actions~\cite{ho2022videodiffusion}, GQN-Mazes, MineRL Navigate, CARLA Town01~\cite{harvey2022flexible}\\
    & &RVD& video production from text &BAIR, KTH Actions, Simulation, Cityscapes~\cite{yang2023diffusion}\\
     &  &DDPM  & 3D object creation & ShapeNet, PartNet~\cite{zhou20213d}\\
     \hline

    Huang et al.~\cite{huang2022draw}& Diffusion Models & MGAD & enhance creativity & Yahoo Flickr Creative Commons 100 million~\cite{thomee2016yfcc100m}\\
    \hline
    Huang et al.~\cite{huang2022diffstyler}& Diffusion Models & DiffStyler & enhance creativity & Conceptual 12M~\cite{changpinyo2021conceptual, crowson2023vdiffusionpytorch} and WikiArt~\cite{wikiart_dataset}\\
    \hline    
    Gregor et al.~\cite{gregor2015draw}& VAE & DRAW & image generation and classification & MNIST~\cite{lecun1998mnist}, Street View House Numbers (SVHN)~\cite{netzer2011reading}, and CIFAR-10~\cite{krizhevsky2009learning}\\  
    \hline
    Xu et al.~\cite{xu2021drb} &  GAN & DRB-GAN & artistic style transfer & WikiArt~\cite{wikiart_dataset} and Place365~\cite{zhou2014learning} \\
    \hline
    Elgammal et al.~\cite{elgammal2017can} &  GAN &  & art generation & WikiArt~\cite{wikiart_dataset} \\    
    \hline
    Kumar et al.~\cite{kumar2016ask} &  VAE & DMN & question answering \\    
    \hline
    OpenAI~\cite{openai2023gpt} &  GAN & GPT-4 &  image-to-text translation & publicly
available data (such as internet data) and data licensed from third-party providers\\ 
    \hline
    Pinaya et al.~\cite{pinaya2023generative} &  Generative Models & MONAI framework &  image-to-image translation & MIMIC-CXR~\cite{johnson2019mimic}, 2D chest X-ray images, CSAW-M~\cite{song2020denoising}, 2D mammograms, UK Biobank~\cite{sudlow2015uk}, 3D T1-
weighted brain images, 2D scenario
with 360,525 extracted 2D brain slices, retinal
optical coherence tomography (OCT)~\cite{kermany2018large} and 2D OCT images \\ 
    \hline
    Güngör et al.~\cite{gungor2023adaptive} & Diffusion Models  & AdaDiff & MRI reconstruction & IXI~\cite{ixi_dataset} and fastMRI~\cite{knoll2020fastmri}\\
    \hline
    Jing et al.~\cite{jing2023eigenfold} &  Diffusion Models &  &  protein structure prediction & PDB IDs~\cite{chakravarty2022alphafold2,saldano2022impact}\\ 
    \hline
    Hu et al.~\cite{hu2023gaia} &  Diffusion Models & GAIA-1 &  self-driving cars & 4,700 hours at 25Hz of proprietary driving data collected in London,
UK between 2019 and 2023\\ 
    \hline
    Antoniou et al.~\cite{antoniou2017data} & GAN & DAGAN &  data augmentation & Omniglot~\cite{lake2015human}, EMNIST~\cite{cohen2017emnist}, and VGG-Faces~\cite{parkhi2015deep} \\ 
    \hline
    Cheng et al.~\cite{cheng2023generative} & VAE & 3D VQ-VAE &  environmental simulation  & MODIS~\cite{giglio2016collection}, VIIRS~\cite{schroeder2014new}\\ 
    \hline
    Nishimura et al.~\cite{nishimura2020human} & GAN & WGAN-GP & gesture generation in robots \\ 
    \hline    
    Katara et al.~\cite{katara2023gen2sim} & Diffusion Models & Gen2Sim & generation of 3D assets, generation of task descriptions, task decomposition & PartNet Mobility~\cite{xiang2020sapien} and GAPartNet~\cite{geng2023gapartnet}\\
    \hline    
    Dong et al. \cite{dong2020fd} & GAN & FD-GAN& Single image dehazing, image restoration & MS COCO \cite{nichol2021glide}, SOTS \cite{li2018benchmarking}, NTIRE'18 \cite{ancuti2018ntire}\\
    \hline
    Yi et al. \cite{yi2024sid} &	GAN	& SID-Net &	Single image dehazing, image restoration	& RESIDE (ITS, OTS, SOTS, HSTS) \cite{li2018benchmarking}, I-Haze \cite{ancuti2018haze}, O-Haze \cite{ancuti2018haze}, Real-world hazy images \cite{choi2015referenceless}\\
    \hline
  \end{tabular}}
  \caption{Applications of different generative models using specific methods and architectures. We use the following abbreviations in the method column: MTGAN (Multi-task GAN), DRB-GAN (Dynamic Resblock GAN), WGAN-GP (Wasserstein GAN with gradient penalty), FD-GAN (Fusion-Discriminator GAN), SID-Net (Single Image Dehazing Network), DAGAN (Data Augmentation GAN), 3D VQ-VAE (Three-dimensional Vector-Quantized VAE), DMN (Dynamic Memory Network), DRAW (Deep Recurrent Attentive Writer), DDPM (Denoising Diffusion Probabilistic Model), ADM (Ablated Diffusion Model), FDM (Flexible Diffusion Model), RVD (Residual Video Diffusion), MGAD (Multimodal Guided Artwork Diffusion), AdaDiff (Adaptive Diffusion Priors).}
  \label{tab:application}
\end{table*}

\begin{figure}[!htb]
  \centering
  \includegraphics[height=10cm,width=14cm]{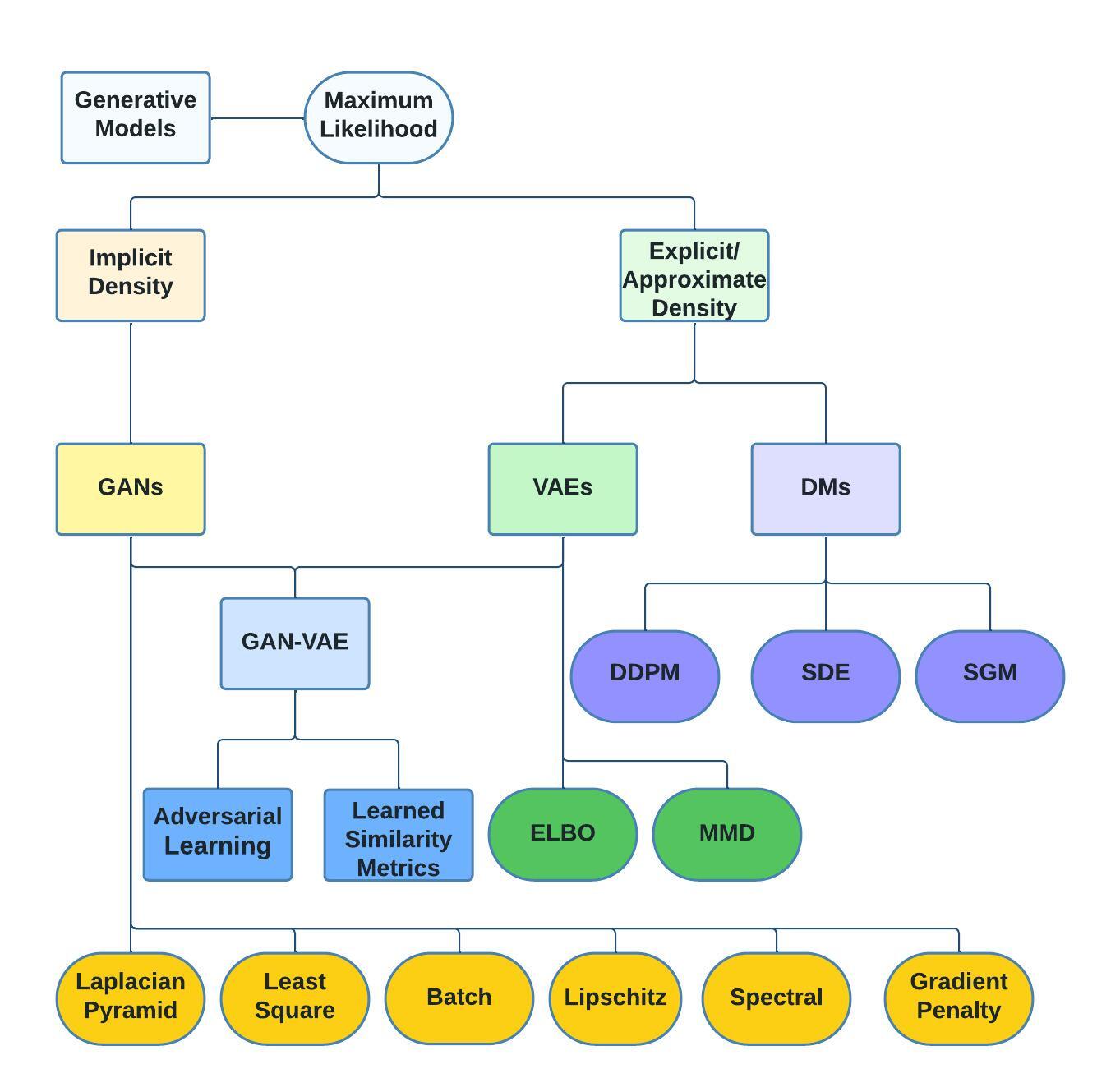}
  \caption{We categorize generative models into two branches based on Likelihood approaches. GANs use implicit density and are classified into six branches based on two perspectives: (1) loss functions and regularization techniques, like batch and spectral normalization, enforcement of Lipschitz continuity, gradient penalty, and least square loss function; and (2) architectures of generators and discriminators: Laplacian pyramid. VAEs use explicit density and are classified into two branches based on use of ELBO and MMD techniques. Hybrid GAN-VAE approaches categorized into two branches based on using adversarial learning and learned similarity metrics. DMs are categorized into three branches based on their main formulations: DDPM, SDE, SGM.
}
  \label{fig:example-taxonomy}
\end{figure}

\section{Taxonomy}
\label{taxonomy}

We organize generative models into two branches based on the fundamental principle of maximum likelihood, distinguishing between implicit and explicit distributions \cite{article}
as presented in Figure \ref{fig:example-taxonomy}. Specifically, \textit{Generative adversarial networks} \cite{goodfellow2014generative}, are direct implicit density models. 
On the other hand, \textit{Variational autoencoders}~\cite{kingma2013auto, rezende2014stochastic} and \textit{Diffusion models}~\cite{ho2020denoising} are approximate density models \cite{bond2021deep}, a type of explicit density models characterized by the explicit definition and maximization of likelihood. 

For GANs, further categorizations are based on techniques applied to augment stability during the training process: Lipschitz constraint~\cite{arjovsky2017wasserstein}, Laplacian pyramid with a conditional GAN~\cite{mao2017least}, gradient penalty~\cite{gulrajani2017improved}, batch normalization~\cite{radford2015unsupervised}, spectral normalization~\cite{miyato2018spectral}, and least square loss function.
Recent advances in VAEs are further categorized into approaches based on Maximum-Mean Discrepancy (MMD) \cite{zhao2017infovae} and Evidence Lower Bound (ELBO) \cite{kingma2013auto}.
VAE and GAN hybrids are also explored. Hybrids are further classified into Adversarially Learned Inference (ALI)~\cite{dumoulin2016adversarially} and autoencoding using learned similarity metrics~\cite{larsen2016autoencoding}, both seamlessly integrating the strengths of VAEs and GANs to enhance generative capabilities.
DMs are categorized based on three main formulations: Denoising Diffusion Probabilistic Models (DDPMs)~\cite{cao2022exploring}, Score-Based Generative Models (SGMs)~\cite{de2021diffusion}, and Stochastic Differential Equations (Score SDEs)~\cite{yang2023diffusion}. 

Building on this taxonomy, Table~\ref{tab:application} provides a comprehensive overview of the generative AI literature focused on image and video synthesis. Each work is summarized across four key dimensions: Type, Method, Datasets, and Applications. The Type refers to the class of generative model (e.g., VAE, GAN, hybrid, or diffusion model), while Method specifies the particular architecture or approach used. Datasets indicate the training data employed, and Applications describe the target tasks or domains addressed by the work. Additionally, we discuss the main strengths and limitations of each model type in Sections~\ref{GANs} (GANs), \ref{VAEs} (VAEs), \ref{Hybrids} (Hybrid models), and \ref{Diffusion} (Diffusion models).

\section{Generative Adversarial Networks (GANs)}
\label{GANs}
GANs \cite{goodfellow2014generative} revolutionized generative AI, enabling unconditional synthesis of visual content by estimating data's underlying probability distribution, overcoming various challenges. In this section, we present a thorough analysis of the architecture and training process of GANs. That is followed by a comprehensive overview of proposed variants based on the technique employed (batch normalization, spectral normalization, laplacian pyramid, lipschitz penalty, gradient penalty, and least square loss function) to improve on the issues faced by traditional GANs, and finally, the challenges that persist. 

\subsection{GAN Architecture}

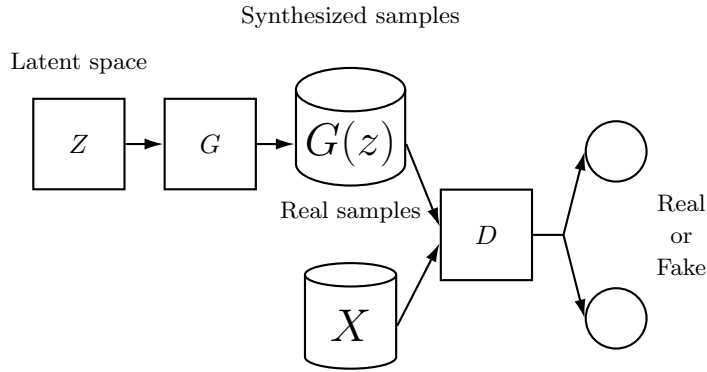
\begin{figure}[h]
    \centering
    \begin{tikzpicture}[
    node distance=1.5cm and .5cm, 
    block/.style={rectangle, draw, thick, minimum width=1.2cm, minimum height=1.2cm, text centered},
    block2/.style={rectangle, draw, thick, minimum width=1.2cm, minimum height=3.2cm, text centered},
    database/.style={cylinder, shape border rotate=90, draw, thick, minimum height=1.4cm, minimum width=1.2cm, aspect=0.4, text centered, font=\Large},
    decision/.style={circle, draw, thick, minimum size=0.8cm},
    arrow/.style={-{Latex[length=2.5mm, width=1.5mm]}, thick} 
  ]

  \node (Z) [block, label={[yshift=0.2cm]above:\small{Latent space}}] {$Z$};
  \node (G) [block, right=of Z] {$G$};
  \node (Gz) [database, right=of G, label={[yshift=0.4cm]above:\small{Synthesized samples}}] {$G(z)$};
  \node (X) [database, below=1cm of Gz, label={[yshift=0.4cm]above:\small{Real samples}}] {$X$};

  \node[block, right=0.5cm of {$(Gz.east)!0.5!(X.east)$}] (D) {$D$};
  \node[decision, above right=0.2cm and 0.8cm of D] (out1) {};
  \node[decision, below right=0.2cm and 0.8cm of D] (out2) {};
  \node[align=center, right=1.5cm of D, yshift=0cm] (outlabel) {\small Real \\ \small or \\ \small Fake}; 

  \draw[arrow] (Z) -- (G);
  \draw[arrow] (G) -- (Gz);

  \draw[arrow] (Gz.east) -- (D.170); 
  \draw[arrow] (X.east) -- (D.190);  

  \coordinate (D_branch) at ($(D.east) + (0.4cm, 0)$); 
  \draw[thick] (D.east) -- (D_branch); 
  \draw[arrow] (D_branch) -- (out1.west);
  \draw[arrow] (D_branch) -- (out2.west);
\end{tikzpicture}
 \caption{Schematic illustration of the Generative Adversarial Network (GAN) architecture, showcasing the interplay between the generator $(G)$, which synthesizes new data samples from latent space $(Z)$, and the discriminator (D), which distinguishes between real data $(X)$ and generated samples $(G(z))$.} 
 \label{fig:figure-gans}
\end{figure}

The architecture of a \textit{Generative Adversarial Network (GAN)} consists of two neural networks: a \textit{generator (G)} and a \textit{discriminator (D)}, which are trained simultaneously in a competitive, adversarial setting~\cite{goodfellow2014generative}. As illustrated in Figure~\ref{fig:figure-gans}, the generator learns to map samples from a prior noise distribution \( z \sim p_z(z) \) to the data space, aiming to produce outputs that resemble real data. Conversely, the discriminator is trained to distinguish between real samples drawn from the true data distribution \( p_{\text{data}}(x) \) and fake samples generated by the generator.

The generator and discriminator engage in a \textit{two-player minimax game}, where the generator tries to fool the discriminator with realistic outputs, while the discriminator improves its ability to classify inputs correctly as real or generated. This interaction is captured by the following value function \( V(D, G) \):

\begin{equation}
\min_G \max_D V(D,G) = \mathbb{E}_{x \sim p_{\text{data}}(x)} [\log D(x)] + \mathbb{E}_{z \sim p_{z}(z)} [\log(1 - D(G(z)))]
\label{eq:gan-objective}
\end{equation}
This adversarial training framework is central to the success of GANs, encouraging the generator to improve the fidelity and realism of its outputs with each iteration. Ideally, as training progresses, the generator becomes so effective that the discriminator can no longer reliably distinguish between real and generated data.

\subsection{GAN Training}
The minimax game between the generator and discriminator networks highlights the most crucial aspect in training GAN models: ensuring that the generator and the discriminator are on par. If either outperforms the other, the training will become unstable, and no useful information will be learned~\cite{goodfellow2014generative}. As a result, researchers have conducted various studies to help ensure this. Arjovsky and Bottou \cite{arjovsky2017wasserstein} made significant contributions by developing principled methods and analysis tools to sidestep the instability issues encountered during the training of GANs. 

Other challenges may also emerge during the training of GANs, 
including issues like divergence, mode collapse \cite{goodfellow2016nips,aro_ge_lia-17a}, vanishing gradients \cite{goodfellow2016nips}, and unstable training~\cite{sajeeda2022exploring}. 
Researchers have proposed various approaches to tackle these hurdles. Divergence becomes a concern when training GANs with gradient descent techniques. In the past, suggestions were made to find a Nash equilibrium \cite{ratliff2013characterization} as a solution, but this proved to be a challenging endeavor~\cite{jabbar2021survey}. 
To address the risk of the generator overtraining the discriminator, techniques like feature matching~\cite{salimans2016improved} have been introduced. This involves setting a non-saturating objective to prevent undesirable outcomes. To combat mode collapse in the generator, a technique called minibatch discrimination~\cite{radford2015unsupervised} has been developed. This involves batching multiple data examples and examining them in combination, which not only helps prevent mode collapse but also enhances the overall stability of the training process. However, batching introduces intra-batch dependencies, which may hinder generalization. To tackle this, virtual batch normalization~\cite{salimans2016improved} has been proposed as a solution. This technique improves the optimization of the neural network by using a reference batch of data examples, reducing statistical dependency and addressing the raised concerns. Furthermore, enhancing the robustness of neural networks against adversarial examples involves strategies such as using historical averages and label smoothing~\cite{salimans2016improved}. Collectively, these techniques contribute significantly to improving the stability of GAN training and the generation of more realistic samples.

Despite the advancements in synthetic sample generation achieved by traditional GANs, challenges like unstable training and vanishing gradients persist. While the aforementioned techniques help improve performance, overcoming these obstacles has yet to be achieved.

\subsection{GAN Variants and Improvements}
\label{GAN-variants}

In the realm of Generative Adversarial Networks (GANs), researchers have actively explored numerous architectural innovations to overcome fundamental training challenges such as instability, vanishing gradients, and mode collapse. This section highlights key developments aimed at enhancing training stability and boosting performance. 
One notable advancement is the introduction of Batch Normalization, a technique that normalizes layer inputs to have zero mean and unit variance. This mitigates internal covariate shift, thereby stabilizing and accelerating the training process~\cite{ioffe2015batch}. The incorporation of Batch Normalization proved especially impactful in the development of the Deep Convolutional GAN (DCGAN)~\cite{radford2015unsupervised}, which marked a significant step forward in stabilizing GAN training and improving the quality of generated outputs. 

DCGAN introduced several crucial architectural modifications to conventional convolutional neural networks (CNNs) tailored to GANs. First, it replaced spatial pooling layers with strided convolutions in the discriminator and fractionally-strided (transposed) convolutions in the generator, allowing the networks to learn up-sampling and down-sampling operations directly. Second, the removal of fully connected hidden layers in deeper architectures reduced overfitting and enabled more efficient learning. Most importantly, Batch Normalization was applied to both the generator and discriminator 
which significantly helped in stabilizing training dynamics and avoiding issues like vanishing gradients. The success of DCGAN has extended its influence across various creative applications, including image synthesis, style transfer, and data augmentation. Although there persist challenges like mode collapse, where the generator produces limited modes of the data distribution, DCGAN has played a foundational role in advancing generative modeling. It not only demonstrated how architectural refinements can stabilize adversarial training but also inspired a broad range of subsequent work in deep generative modeling.

To address the issue of mode collapse in GANs, a significant advancement came with the introduction of the Wasserstein GAN (WGAN) \cite{arjovsky2017wasserstein}. Unlike traditional GANs that use a discriminator to classify real vs. fake samples, WGAN replaces this with a critic network that estimates the Wasserstein (or Earth Mover’s) distance between the real and generated data distributions. This metric provides smoother gradients and a more meaningful loss function, leading to more stable training. However, the original WGAN enforces the Lipschitz continuity condition (required for the Wasserstein distance to be valid) using weight clipping, which can result in optimization difficulties and poor-quality samples. To overcome these limitations, WGAN-GP (WGAN with Gradient Penalty) \cite{gulrajani2017improved} was proposed. Instead of clipping weights, it adds a gradient penalty term to the loss function, which constrains the norm of the gradient of the critic with respect to its input. This soft constraint more effectively enforces Lipschitz continuity and improves convergence and sample quality. Together, these innovations, WGAN and WGAN-GP, have substantially improved the stability of GAN training and enhanced the quality and diversity of generated samples.

Another class of GAN variants addresses unstable training through a technique known as spectral normalization, as introduced in Spectral Normalization GANs (SNGANs) \cite{miyato2018spectral}. In this approach, the authors propose a novel weight normalization method that directly controls the Lipschitz constant of the discriminator by normalizing the spectral norm (i.e., the largest singular value) of each weight matrix. By enforcing this constraint, SNGANs stabilize the training dynamics of the discriminator, which in turn leads to more reliable gradients for the generator. A key advantage of spectral normalization is that it reduces the need for extensive hyperparameter tuning typically associated with enforcing Lipschitz continuity (as in WGAN-GP), thereby lowering computational overhead. This technique not only improves training stability but also enhances the quality and realism of the generated outputs, making it a practical and efficient solution for robust GAN training.

One of the key challenges in training traditional GANs is the vanishing gradient problem, which occurs when the discriminator becomes overly confident and provides minimal feedback to the generator. To mitigate this, Least Squares GANs (LSGANs) \cite{mao2017least} propose replacing the standard binary cross-entropy loss with a least squares loss. The key idea behind this change is to shape the loss landscape in a way that provides stronger gradients even when the discriminator is confident, thereby facilitating better learning for the generator.

In traditional GANs, when the discriminator confidently classifies a fake sample as fake (i.e., output close to 0), the gradient passed back to the generator becomes very small. This causes the generator to stop learning effectively, leading to vanishing gradients. LSGANs address this by treating the discriminator's output as a regression target rather than a classification probability. The least squares loss penalizes the difference between the discriminator's output and a target value (e.g., 1 for real, 0 for fake) using a mean squared error formulation. This results in a smoother and more informative gradient signal, even when the discriminator is performing well. As a consequence, the generator receives meaningful feedback throughout the training, which improves convergence and helps it generate more realistic samples. 

Unlike the original GAN formulation that minimizes the Jensen-Shannon divergence, LSGANs effectively minimize the Pearson \(\chi^2\) divergence, resulting in more stable training. This approach also leads to more realistic sample generation and reduces mode collapse, as the generator is guided more directly toward producing outputs that the discriminator cannot distinguish from real data. The objective functions for LSGANs are defined as:

\begin{equation}
\min_D V_{\text{LSGAN}}(D) = \frac{1}{2} \mathbb{E}_{x \sim p_{\text{data}}(x)} \left[(D(x)-b)^2\right] + \frac{1}{2} \mathbb{E}_{z \sim p_{\text{z}}(z)} \left[(D(G(z))-a)^2\right]
\end{equation}

\begin{equation}
\min_G V_{\text{LSGAN}}(G) = \frac{1}{2} \mathbb{E}_{z \sim p_{\text{z}}(z)} \left[(D(G(z))-c)^2\right]
\end{equation}

Here, \( a \) and \( b \) represent the target labels for fake and real data in the discriminator, respectively, while \( c \) is the value that the generator wants the discriminator to assign to fake samples (typically set to \( b \)).

Another significant advancement in GAN research is the Laplacian Pyramid of Adversarial Networks (LAPGAN)~\cite{denton2015deep}, which integrates the concept of conditional GANs (cGANs)~\cite{mirza2014conditional} with a Laplacian pyramid decomposition to facilitate high-quality image generation in a coarse-to-fine manner. While cGANs extend the standard GAN framework by conditioning both the generator and discriminator on auxiliary information \( y \) (such as class labels or other contextual data), LAPGAN leverages this conditioning within a multi-scale architecture to improve the fidelity and realism of generated images.

LAPGAN employs a hierarchy of generative models \(\{G_0, G_1, \ldots, G_K\}\), where each \(G_k\) is trained to model the distribution of residual image details (high-frequency components) at level \(k\) of the Laplacian pyramid. Image synthesis begins at the coarsest level, where the model \(G_K\) generates a low-resolution image. Subsequently, at each finer level \(k < K\), the generator \(G_k\) produces a high-frequency residual conditioned on an upsampled version of the image generated at level \(k+1\). These residuals are then combined with the upsampled images to iteratively reconstruct a high-resolution image. Each generative model \(G_k\) is trained independently using the cGAN framework. During training, Laplacian pyramids are constructed from real images, and the generators learn to replicate the residual distributions at each level. While the primary training procedure is unsupervised, LAPGAN can be adapted to a class-conditional setting by appending a one-hot encoded class vector to both the generator and discriminator inputs. The key innovation of LAPGAN lies in decomposing the generation task into a sequence of conditioned refinements at multiple scales. This approach not only improves training stability but also enhances the visual quality of generated samples, as it allows the model to focus on learning plausible details at each resolution level rather than modeling the full image distribution in a single pass.

Beyond architectural improvements aimed at general image synthesis quality and training stability, the GAN framework has also led to the development of specialized variants tailored for specific image processing and restoration tasks. These variants often adapt the core generator-discriminator architecture with domain-specific knowledge or novel components to tackle complex challenges. For instance, in the field of single image dehazing, FD-GAN~\cite{dong2020fd} introduces a significant modification to the discriminator, known as the  Fusion-Discriminator. This discriminator incorporates both high- and low-frequency components as additional priors, enabling it to provide more nuanced feedback to the generator. This, in turn, allows the generator to produce dehazed images that are not only clearer but also more natural and realistic, with fewer artifacts and better color fidelity compared to those guided by traditional discriminators.
Another innovative variant in this domain is SID-Net~\cite{yi2024sid}, which extends the adversarial learning paradigm by integrating it with contrastive learning. This hybrid approach allows the network to more effectively utilize information from both clean ground truth images (positive samples) and the input hazy images (negative samples). The adversarial component guides the image dehazing process towards realistic, haze-free outputs, while the contrastive learning aspect refines the feature representations to better distinguish between hazy and clean characteristics. Such specialized GAN variants demonstrate the adaptability of adversarial principles, where tailored architectures and learning objectives lead to state-of-the-art performance in complex image-to-image translation tasks.

Recent developments in GAN research have significantly broadened the scope of generative modeling, leading to improved performance and enabling a range of new applications. Nevertheless, persistent issues such as training instability, vanishing gradients, and the complexity of hyperparameter tuning continue to pose challenges.

\subsection{Challenges and Limitations}

Despite advancements in GANs and efforts to address the challenges, which involve modifying architectures, emphasizing regularization, and understanding stability criteria like Lipschitz Continuity, they still face issues such as training convergence problems, mode collapse, and vanishing gradients. The sensitivity of the discriminator to data distribution raises concerns about adversarial attacks affecting GAN performance. While adversarial training enhances robustness, current defense mechanisms are not foolproof, creating vulnerabilities to varied attacks. The absence of a definitive defense mechanism capable of handling diverse attacks remains a challenge~\cite{sajeeda2022exploring}. Additionally, GANs have limitations including high computational and memory requirements, as well as difficulties in evaluating their performance~\cite{kumar2022school}. These limitations can be overcome with further research and advancements in the field of GANs. Avenues for future research include extending GAN robustness beyond gradient-based attacks and enhancing overall training stability.

In conclusion, while GANs have achieved remarkable success in generating high-quality samples, they are not exempt from challenges that continue to impact their performance. Solving the critical issues of unstable training, mode collapse, mode dropping, and efficiently selecting appropriate hyperparameters remain open problems for researchers to focus on.

\section{Variational Autoencoders (VAEs)}
\label{VAEs}
VAEs combine the power of autoencoders and probabilistic modeling to learn meaningful representations of data in a latent space. Instead of constructing an encoder that produces a single value for each latent attribute, the VAE's encoder is designed to describe a probability distribution for each latent attribute. This enables smooth interpolation, meaningful sampling, and high-quality image generation, making VAEs a valuable tool in generative AI.
Here we present a thorough analysis of the architecture and training process of VAEs, followed by a comprehensive overview of variants based on the technique employed (MMD, ELBO)
to improve on the issues faced by traditional VAEs, and the challenges that  persist. 

\subsection{VAE Architecture}\label{AA}
Introduced by Kingma and Welling \cite{kingma2013auto}, the VAE architecture consists of two main components: an encoder (or recognition network) and a decoder (or generator network). The encoder's primary task is to approximate the intractable posterior distribution of latent variables. In contrast, the decoder's role is to generate new samples using the learned latent space representation produced by the encoder. The encoder achieves its goal by maximizing the evidence lower bound (ELBO)~\cite{kingma2013auto}, which balances the reconstruction accuracy of the data and the divergence between the encoder's approximate distribution and the prior distribution of the latent variables. Mathematically, ELBO can be expressed as:

\begin{equation}
\text{ELBO} = E[\log p(x|z)] - \text{KL}(q(z|x) || p(z))
\end{equation}

\noindent where $E[\log p(x|z)]$ represents the reconstruction term, which measures how well the decoder can reconstruct the input data given the latent variables $z$. $\text{KL}(q(z|x) || p(z))$
is the Kullback-Leibler (KL) divergence term, which measures the difference between the encoder's approximate distribution $q(z|x)$  and the prior distribution $p(z)$ of the latent variables. Maximizing ELBO effectively encourages the encoder to learn a meaningful representation of input data in the latent space. 

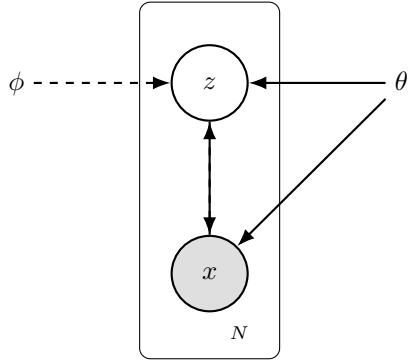
\begin{figure}[ht]
    \centering
    \begin{tikzpicture}[
    >=Latex, 
    node distance=1.5cm and 1.8cm 
  ]

  \tikzstyle{latent} = [circle, draw, thick, minimum size=10mm, inner sep=0pt]
  \tikzstyle{obs}    = [latent, fill=gray!25] 
  \tikzstyle{param}  = [text height=1.5ex, text depth=.25ex] 

  \node[latent]             (z) {$z$};
  \node[obs, below=of z]    (x) {$x$};
  \node[param, left=of z]   (phi) {$\phi$};
  \node[param, right=of z]  (theta) {$\theta$};

  \edge[thick] {theta} {z};   
  \edge[thick] {z} {x};       
  \edge[thick] {theta} {x};   

  \edge[thick, dashed] {phi} {z};   
  \edge[thick, dashed, bend left=35] {x} {z}; 

  \plate[inner sep=0.4cm, xshift=-0.0cm, yshift=0.15cm] {plate} {(x)(z)} {$N$}; 

\end{tikzpicture}
    \caption{The type of directed graphical model under consideration. Solid lines denote the generative model $p_{\theta}(z)p_{\theta}(x|z)$, dashed lines denote variational approximation $q_{\phi}(z|x)$ to the intractable posterior $p_{\theta}(z|x)$. The variational parameters $\phi$ are learned jointly with the generative model parameter $\theta$.
}
    \label{fig: VAE}
\end{figure}

The encoder consists of multiple layers of neural networks to capture complex relationships in data and map it to the lower-dimensional latent space. Similarly, the decoder comprises neural network layers to upsample the latent variables and reconstruct data from the latent space representation. By jointly training the encoder and decoder, VAE learns to generate new data points resembling the original data distribution. Illustrating this two-part structure, Figure~\ref{fig: VAE} shows a directed graphical model representing the encoder’s variational approximation \( q_{\phi}(z|x) \), and the decoder’s generative model \( p_{\theta}(z)\,p_{\theta}(x|z) \), where \( \phi \) denotes the parameters of the approximate posterior distribution over the latent variables, and \( \theta \) represents the parameters of the generative model.
 \par

Another type of VAE architecture integrates the deep neural network and approximate Bayesian inference to learn the latent space distribution and generate new realistic samples of data~\cite{rezende2014stochastic}. With this architecture, the recognition model acts as a stochastic encoder to approximate the representation of the posterior over the latent variables, and the generative model uses an objective function based on the variational principle for optimizing parameters. This approach allows training the model by gradient backpropagation for jointly optimizing the parameters of the generative and recognition models.

\subsection{VAE Training}\label{AA}

Kingma and Welling~\cite{kingma2013auto} proposed a seminal framework for unsupervised learning using \emph{Variational Autoencoders (VAEs)} with continuous latent variables. The core objective of this method is to approximate the intractable \emph{marginal likelihood} of the data \( \log p_{\theta}(X) \), where \( X \) denotes the observed data. Direct maximization of this marginal likelihood is generally infeasible due to the difficulty of integrating over the latent variables. Instead, their approach focuses on maximizing a \emph{variational lower bound} \( \mathcal{L}(\theta, \phi; X) \), also known as the \emph{evidence lower bound (ELBO)}.

Traditional methods such as mean-field Variational Bayes (VB) or Expectation-Maximization (EM) can be computationally expensive in this setting. To address this, Kingma and Welling introduced the \emph{reparameterization trick}, which expresses a latent variable \( z \sim q_{\phi}(z|x) \) as a differentiable transformation of a noise variable \( \epsilon \sim p(\epsilon) \), typically standard normal, through a function \( z = g_{\phi}(x, \epsilon) \). This allows gradients to propagate through stochastic nodes in the computational graph, enabling optimization of the ELBO using \emph{gradient-based stochastic optimization} methods such as Adam.

The training objective includes a \emph{Maximum A Posteriori (MAP)} estimate of the parameters \( \theta \) and \( \phi \), with a Gaussian prior \( p(\theta, \phi) = \mathcal{N}(0, I) \) imposed for regularization. Under this formulation, MAP estimation corresponds to maximizing the log-likelihood with an additional \emph{weight decay} term, penalizing large parameter values.

To make the optimization scalable, especially for large datasets, the gradient of the log marginal likelihood is approximated by the gradient of the variational lower bound:
\begin{equation}
    \nabla_{\theta,\phi} \log p_{\theta}(X) \approx \nabla_{\theta,\phi} \mathcal{L}(\theta,\phi; X)
    \label{eq: max_likelihood}
\end{equation}
This approximation forms the basis for efficiently training VAEs, making them a practical tool for unsupervised learning with deep generative models.


Furthermore, backpropagation enables the computation of gradients, which are then used to update the parameters of both the encoder and decoder components during training~\cite{kingma2013auto}. Using the encoder network, the VAE can effectively learn to encode input data into a lower-dimensional latent space. 
Subsequently, the decoder network generates new samples from the learned latent space. 

Building upon the foundational work of Kingma and Welling~\cite{kingma2013auto}, a novel training paradigm combines gradient-based backpropagation with approximate inference to improve computational efficiency and scalability in deep generative models. This approach optimizes both the generative model parameters and the variational parameters by leveraging the reparameterization trick, which enables gradients to be computed through stochastic latent variables modeled using Gaussian distributions. Rather than computing gradients over the entire dataset, the training uses mini-batches, significantly reducing computational overhead and allowing the method to scale to large datasets.

An important contribution to scalable variational inference is the use of stochastic optimization techniques, such as those introduced by Hoffman et al.~\cite{hoffman2013stochastic}, which involve subsampling data to compute noisy estimates of the gradient. While their method was initially proposed for models like Latent Dirichlet Allocation, the idea of combining stochastic optimization with variational inference has strongly influenced the design of efficient VAE training algorithms. These advancements have significantly improved the practicality of training VAEs, enabling them to model complex data distributions, generate high-quality samples, and handle large-scale datasets with continuous latent spaces. The combination of the reparameterization trick, backpropagation, and stochastic optimization forms the foundation of the modern VAE framework, which has seen widespread success in applications such as image generation, representation learning, and data compression.

Furthermore, techniques such as incorporating auxiliary variables and employing learned similarity metrics have been explored in various VAE extensions to improve representational capacity and sample quality. Despite these improvements, VAEs still face challenges, such as producing blurry outputs, especially when modeling multimodal or high-complexity data. Numerous variants have been proposed to mitigate these issues, as discussed in Section~\ref{VAE-variants}.


\subsection{VAE Variants and Improvements}
\label{VAE-variants}

A known limitation in the training of Variational Autoencoders (VAEs) is the problem of \textit{posterior collapse}, where the decoder learns to ignore the latent variables and relies solely on the autoregressive nature or capacity of the decoder network. This often results in a poor utilization of the latent space, particularly when using powerful decoders such as RNNs or deep CNNs, leading to low diversity in generated outputs. To address this issue, several improved training objectives have been proposed. One notable direction involves modifying the Evidence Lower Bound (ELBO) to encourage better alignment between the approximate posterior and the prior, while still preserving useful information in the latent code. A prominent example of this is the InfoVAE framework \cite{zhao2017infovae}, which generalizes the ELBO by incorporating information-theoretic terms that better control the trade-off between reconstruction quality and latent space regularization. Within this framework, the Maximum Mean Discrepancy VAE (MMD-VAE) is a specific variant that replaces the traditional Kullback–Leibler (KL) divergence term with the Maximum Mean Discrepancy (MMD) \cite{gretton2008kernel,li2015generative, dziugaite2015training}. MMD is a non-parametric metric that quantifies the distance between two probability distributions based on the differences in all their moments. This is achieved using kernel methods, which enable efficient computation via the kernel trick. Given a positive definite kernel function \( k(\cdot, \cdot) \), the squared MMD between two distributions \( p \) and \( q \) is defined as:

\begin{equation}
\begin{aligned}
    D_{\text{MMD}}(q \| p) = \mathbb{E}_{z, z' \sim p(z)}[k(z, z')] - 2\mathbb{E}_{z \sim q(z), z' \sim p(z')} [k(z, z')] + \mathbb{E}_{z, z' \sim q(z)} [k(z, z')]
\end{aligned}
\label{eq:MMD}
\end{equation}

The MMD is zero if and only if \( p = q \), making it a powerful tool for distribution matching in generative models. In the context of VAEs, using MMD as a regularization term helps encourage the aggregated posterior \( q(z) \) to match the prior \( p(z) \), while avoiding the collapse of the latent representation. This leads to richer latent encodings and more diverse and meaningful generative outputs.

\par

Variational Autoencoders (VAEs) have undergone numerous enhancements and architectural modifications aimed at overcoming their initial limitations and improving their ability to generate more realistic and accurate image samples. A significant advancement involves the use of learned similarity metrics, which shift the reconstruction objective away from simple pixel-wise errors (like mean squared error) towards perceptually meaningful distances. By aligning reconstructions with features extracted from pretrained neural networks (e.g., VGG networks), these models promote the learning of more informative latent space representations and higher-quality outputs.

Another important line of improvement is the incorporation of auxiliary variables, as proposed in the Auxiliary Deep Generative Model (ADGM) by Maaløe et al.~\cite{maaloe2016auxiliary}. In this semi-supervised VAE variant, auxiliary variables are introduced to enrich the expressiveness of the variational posterior distribution. This results in a better approximation of the true posterior and improves the model’s ability to capture complex data dependencies. Techniques such as temperature scaling, hierarchical latent structures, and multiple layers of stochastic variables are often integrated to further boost the flexibility and robustness of the model.

Despite these advancements, several challenges persist. Notably, posterior collapse where the decoder disregards latent variables and difficulties in modeling highly complex or multimodal distributions continue to pose barriers to the wider adoption of VAEs. Nonetheless, each new variant contributes to refining the framework and expanding its practical utility in domains like image generation, semi-supervised learning, and representation learning.

\subsection{Challenges and Limitations}
\label{VAE-challenges}
Exploring the challenges and limitations still associated with VAEs and their variants is crucial for further advancements in the field. While MMD-VAE improves on the traditional VAE architecture to counter the issue of posterior collapse, they are unable to address it definitively. Similarly, while semi-supervised models improve upon traditional VAEs and help capture more complicated variational distributions, the problem is by no means solved. Each of these directions represent open problems for researchers to focus on to help unlock the full potential of VAEs in the domain of generative AI. To address the prevailing limitations of GANs and VAEs, researchers have introduced hybrid models to amalgamate their strengths for a boost in performance as detailed in Section \ref{Hybrids}.

\section{Diffusion Models (DMs)}
\label{Diffusion}

Diffusion Models (DMs) \cite{ho2020denoising,croitoru2023diffusion} have emerged as a transformative class of deep generative models, outperforming prior approaches like GANs in many tasks, particularly in image synthesis \cite{ho2020denoising}, video generation~\cite{ho2022videodiffusion}, and molecular design~\cite{yang2023diffusion,xu2022geodiff, yang2024graphusion}. The core idea behind DMs is to progressively add noise to data in a forward process and then train a model to reverse this process, thereby generating new data samples from noise. DMs typically operate in two phases: 

\noindent\textit{Forward diffusion process:} This is a predefined, usually Gaussian-based, Markov chain \cite{ho2020denoising,sohl2015deep} that gradually corrupts the input data over \(T\) steps, transforming it into pure noise.

\noindent\textit{Reverse denoising process:} A parameterized neural network is trained to reverse the corruption process, effectively learning to generate realistic data from noise~\cite{ho2020denoising, song2020score}. 

In this section, we elaborate on the architectural components that power DMs, focusing on the design of denoising networks, the role of different architectures like U-Net and Transformer, and how hybrid approaches leverage the strengths of both.

\subsection{Diffusion Models Architecture}
At the core of a diffusion model lies a critical component known as the denoising network. This network is responsible for guiding the reverse diffusion process, which gradually transforms random noise back into meaningful, structured data such as images, text, or graphs. During training, diffusion models learn to simulate the step-by-step corruption of data by adding noise, and the denoising network learns how to reverse this process.

The denoising network’s main task at each timestep is to predict either the original, clean data sample or the specific amount of noise that was added at that step. In doing so, it essentially learns to approximate the probability distribution of what the data should look like one step earlier, given its current noisy state. Mathematically, this involves estimating the reverse conditional distribution \( p_\theta(x_{t-1} | x_t) \), which represents the probability of obtaining the data sample from the previous timestep \(x_t\),  given the current noisy version \(x_t\). Here, \(\theta\) represents the set of parameters (i.e., the weights) of the denoising neural network, which are learned during training. Alternatively, the score function \( \nabla_x \log p(x_t) \), which represents the gradient of the log-probability of the data at timestep \(t\) and provides direction on how to denoise the input efficiently. These two formulations are central to modern diffusion modeling. The reverse conditional distribution is typically used in denoising diffusion probabilistic models (DDPMs), while the score function arises from score-based generative modeling frameworks, as introduced by Song et al.~\cite{song2020score}. More recently, these ideas have been adapted to structured data types like graphs, as demonstrated in the Graphusion framework by Yang et al.~\cite{yang2024graphusion}. To implement the denoising function, diffusion models rely on powerful neural network architectures. Two architectural families have become especially prominent: U-Nets~\cite{ronneberger2015u} and Transformers~\cite{vaswani2017attention}.

The U-Net architecture, originally introduced for image segmentation~\cite{ronneberger2015u}, has been widely adopted in diffusion models due to its strong inductive bias for image data and its effective use of multi-scale features. U-Nets follow an encoder-decoder structure: The encoder downsamples the input to extract hierarchical features and the decoder upsamples while merging features from earlier layers via skip connections, allowing for fine-grained reconstructions. In diffusion models, such as DDPMs~\cite{ho2020denoising, sohl2015deep}, this design allows efficient denoising by reusing multi-scale representations. U-Net variants have been further extended with \textit{attention mechanisms}, as seen in Latent Diffusion Models (LDMs)~\cite{rombach2022high}, and with autoregressive components like \textit{PixelCNN++}~\cite{salimans2017pixelcnn++}.

Recently, Transformers~\cite{vaswani2017attention}, known for their success in natural language processing and vision, have been introduced into diffusion models due to their ability to model global dependencies~\cite{chang2023design}. Their self-attention mechanism enables flexible contextual modeling across the entire spatial domain, which is especially beneficial in high-resolution generation tasks. In diffusion settings,  Transformers can be used in both encoder-decoder formats (e.g., Diffuser~\cite{popov2021grad}) and in pure encoder settings with task-specific decoders~\cite{cao2022exploring}. However, Transformers alone may struggle with capturing fine local details. To address this, hybrid architectures that combine \textit{U-Net skip connections} with Transformer blocks have been proposed~\cite{chang2023design}. For instance, U-ViT~\cite{bao2022all} and DiT~\cite{peebles2023scalable} architectures embed Transformer layers within the U-Net framework, maintaining locality through skip connections while leveraging the global modeling power of self-attention.

Transformer-based denoising networks exhibit additional desirable properties compared to U-Net architectures. Specifically, they achieve comparable performance in conditional image generation
~\cite{chahal2022exploring}, and superior quality with reduced network complexity in unconditional image generation tasks. Researchers have explored the use of graph transformers to capture relationships in graph-structured data. Furthermore, Transformers contribute to scalability~\cite{chang2023design}, and multi-modality in DMs.

\subsection{Diffusion Models Training}
Diffusion models (DMs) are a class of latent variable models that generate data by learning to reverse a predefined stochastic process that gradually corrupts data with noise. The generative process in DMs is defined through a latent variable model using a fixed-length Markov chain. Specifically, the data distribution is modeled as:
\begin{equation}
p_\theta(x_0) := \int p_\theta(x_{0:T}) \, dx_{1:T}
\end{equation}
Here, \(x_0\in \mathbb{R^d}\) denotes the observed data, assumed to be drawn from the data distribution \( x_0 \sim q(x_0) \), and \( x_1, x_2, \dots, x_T \) are latent variables of the same dimensionality as \( x_0 \), representing progressively noisier versions of the original data. The joint distribution \( p_\theta(x_{0:T}) \) specifies the reverse process, which is a parameterized Markov chain defined as:
\begin{equation}
\begin{aligned}
p_\theta(x_{0:T}) &:= p(x_T) \prod_{t=1}^T p_\theta(x_{t-1} \mid x_t), \\
p(x_T) &= \mathcal{N}(x_T; 0, I), \\
p_\theta(x_{t-1} \mid x_t) &:= \mathcal{N}(x_{t-1}; \mu_\theta(x_t, t), \Sigma_\theta(x_t, t))
\end{aligned}
\label{eq:DM}
\end{equation}
where \( p(x_T) \) is the prior distribution over the final timestep, typically chosen as a standard multivariate Gaussian:
\[
p(x_T) = \mathcal{N}(x_T; 0, I).
\]
The transition distributions \( p_\theta(x_{t-1} \mid x_t) \) are modeled as Gaussians whose mean \( \mu_\theta(x_t, t) \) and covariance \( \Sigma_\theta(x_t, t) \) are predicted by a neural network with parameters \( \theta \).

Unlike other latent variable models, diffusion models fix the approximate posterior \( q(x_{1:T} \mid x_0) \), also called the forward or diffusion process. This process is defined as a Markov chain that incrementally corrupts the input data by adding Gaussian noise based on a predefined variance schedule \( \beta_1, \dots, \beta_T \), where \( \beta_t \in (0,1) \) controls the noise level at each timestep. The forward process is defined as:

\begin{equation}
\begin{aligned}
q(x_{1:T} \mid x_0) &:= \prod_{t=1}^{T} q(x_t \mid x_{t-1}), \\
q(x_t \mid x_{t-1}) &:= \mathcal{N}(x_t; \sqrt{1 - \beta_t} \, x_{t-1}, \beta_t I),
\end{aligned}
\label{eq:Markovchain}
\end{equation}

where \( I \) denotes the identity matrix. This recursive Gaussian corruption gradually drives \( x_t \) towards a nearly isotropic Gaussian distribution as \( t \to T \).

Training a diffusion model involves minimizing the negative log-likelihood of the data under the model. Since directly optimizing \( -\log p_\theta(x_0) \) is intractable, a variational upper bound is used, resulting in the evidence lower bound (ELBO):

\begin{equation}
\begin{aligned}
\mathbb{E}[-\log p_\theta(x_0)] &\leq \mathbb{E}_q \left[ -\log\left(\frac{p_\theta(x_{0:T})}{q(x_{1:T} \mid x_0)}\right) \right] \\
&= \mathbb{E}_q \left[ -\log p(x_T) - \sum_{t \geq 1} \log\left(\frac{p_\theta(x_{t-1} \mid x_t)}{q(x_t \mid x_{t-1})}\right) \right] \\
&:= L,
\end{aligned}
\label{eq:neg_likelihood}
\end{equation}

where the expectation is taken with respect to the forward process \( q \), and \( L \) denotes the training objective.

The variance schedule \( \{ \beta_t \}_{t=1}^{T} \) can be set manually (e.g., linearly or cosine-spaced) or learned during training. To facilitate computations, we define \( \alpha_t := 1 - \beta_t \) and the cumulative product \( \bar{\alpha}_t := \prod_{s=1}^{t} \alpha_s \). Using these, we can express the marginal distribution \( q(x_t \mid x_0) \) in closed form:

\[
q(x_t \mid x_0) = \mathcal{N}(x_t; \sqrt{\bar{\alpha}_t} x_0, (1 - \bar{\alpha}_t) I).
\]

This allows for efficient training, as \( x_t \) can be sampled directly from \( x_0 \) without simulating the entire chain. The training loss \( L \) is optimized using stochastic gradient descent. To reduce the variance of the gradient estimates, the ELBO can be re-expressed in terms of Kullback-Leibler (KL) divergences:

\begin{equation}
\begin{aligned}
&\mathbb{E}_q \left[ D_{\text{KL}}\left(q(x_T \mid x_0) \parallel p(x_T)\right) \right] + \sum_{t>1} D_{\text{KL}}\left(q(x_{t-1} \mid x_t, x_0) \parallel p_\theta(x_{t-1} \mid x_t)\right) - \log p_\theta(x_0 \mid x_1).
\end{aligned}
\label{eq:optimized}
\end{equation}

Each KL divergence compares a Gaussian from the forward process with one from the learned reverse process. When conditioned on the original data \( x_0 \), the posterior \( q(x_{t-1} \mid x_t, x_0) \) also has a closed-form Gaussian representation:

\begin{equation}
q(x_{t-1} \mid x_t, x_0) = \mathcal{N}(x_{t-1}; \tilde{\mu}_t(x_t, x_0), \tilde{\beta}_t I),
\label{eq:KL_optimized}
\end{equation}

where the mean \( \tilde{\mu}_t \) and variance \( \tilde{\beta}_t \) are defined as:

\begin{equation}
\begin{aligned}
\tilde{\mu}_t(x_t, x_0) &:= \frac{\sqrt{\bar{\alpha}_{t-1}} \beta_t}{1 - \bar{\alpha}_t} x_0 + \frac{\sqrt{\alpha_t} (1 - \bar{\alpha}_{t-1})}{1 - \bar{\alpha}_t} x_t, \\
\tilde{\beta}_t &:= \frac{1 - \bar{\alpha}_{t-1}}{1 - \bar{\alpha}_t} \beta_t.
\end{aligned}
\label{eq:Rao-Blackwellized}
\end{equation}

This structure allows the KL terms in Equation~\ref{eq:optimized} to be computed in closed form, a technique often referred to as Rao-Blackwellization. This approach avoids high-variance Monte Carlo estimation and was originally employed by Ho et al.~\cite{ho2020denoising}.

Diffusion models offer several advantages over adversarial generative models like GANs. While GANs suffer from instability due to the adversarial training objective and are prone to mode collapse~\cite{salimans2016improved}, DMs benefit from a likelihood-based formulation that yields stable training and enhanced sample diversity. However, diffusion models are computationally less efficient during inference, as they require multiple sequential passes through the neural network. In contrast to GANs, which employ a compact latent space (often \( \mathbb{R}^{100} \)), diffusion models retain the full dimensionality of the data (e.g., images in \( \mathbb{R}^{H \times W \times C} \)). The latent variables in DMs follow a standard Gaussian prior, similar to variational autoencoders (VAEs).

Recent work has shown that GANs’ latent spaces often encode semantically meaningful directions, enabling structured manipulations of generated images. In contrast, the latent space in DMs is less interpretable but more expressive~\cite{croitoru2023diffusion}. As we move forward, we explore architectural and algorithmic extensions to diffusion models aimed at addressing these limitations while retaining their strengths.

\subsection{Diffusion Models Variants and Improvements}

De-noised Diffusion Probabilistic Models (DDPMs)~\cite{cao2022exploring} are foundational to the modern development of diffusion-based generative models. DDPMs formulate data generation as a two-stage process: a forward diffusion process that gradually adds noise to the data, and a reverse denoising process that reconstructs the data from noise. In the \textit{forward process}, DDPMs define a sequence of transition kernels by selecting a noise schedule such as constant, linear, or cosine schedules~\cite{nichol2021improved} to determine how Gaussian noise is progressively added over \( T \) timesteps. This noise addition forms a Markov chain that incrementally corrupts the input data, driving it toward an isotropic Gaussian distribution. These diffusion steps are inspired by thermodynamic processes, such as heat dissipation, modeling a gradual loss of information in a controlled manner. The \textit{reverse process} is modeled using learnable Gaussian transitions parameterized by \( \theta \). The objective is to approximate the true reverse-time dynamics of the forward process and recover data samples by sequentially denoising the noisy inputs.

The training objective of DDPM is to minimize a variational bound on the negative log-likelihood (NLL), similar to VAEs. The objective consists of a prior loss ($\mathcal{L}_T$), a reconstruction loss ($\mathcal{L}_0$), and the sum of divergences between the posterior of forward steps and the corresponding reverse steps ($\mathcal{L}_{1:T-1}$). The model is trained to minimize the negative log-likelihood by focusing on the $\mathcal{L}_{1:T-1}$ term. The diffusion training objective involves the expectation of an $\ell_2$-loss between mean coefficients in the denoising process. This is expressed through a simplified training objective, termed $\mathcal{L}_{\text{simple}}$, which incorporates noise schedules and denoising score-matching objectives. After training, the neural prediction is used in the reverse process for ancestral sampling.

In practice, the DDPM objective is often simplified into a noise prediction task. The resulting simplified loss, denoted as \( \mathcal{L}_{\text{simple}} \), minimizes the mean squared error (MSE) between the true noise and the predicted noise added to the data at each timestep.
This objective is equivalent to denoising score matching under certain assumptions~\cite{vincent2011connection}. The model learns to estimate the noise component at each step and uses this information to perform ancestral sampling in the reverse direction.

While many diffusion models adopt the DDPM framework, several important variants and improvements have been proposed. One prominent extension is the Improved DDPM~\cite{nichol2021improved}, which augments \( \mathcal{L}_{\text{simple}} \) with auxiliary losses to improve sample quality and training stability. These include: A learned noise schedule to replace the fixed beta schedules, a variational lower bound loss on the signal-to-noise ratio, and hybrid losses combining noise prediction, data prediction, and perceptual metrics. Despite these enhancements, the \( \mathcal{L}_{\text{simple}} \) formulation remains a fundamental component across most DDPM-based models due to its simplicity, efficiency, and empirical success.


Score-based generative models form a distinct class within the diffusion model framework, grounded in both theoretical foundations and empirical advancements. These models aim to directly estimate the \textit{score function}, i.e., the gradient of the log-density of noisy data, to enable sample generation through score-based sampling techniques~\cite{song2020improved}. During training, innovative strategies have been proposed for choosing noise scales and conditioning neural networks on these noise levels. In particular, Song and Ermon~\cite{song2019generative} introduced \textit{Non-Compositional Stochastic Networks} (NCSNs), which learn the score function using a noise-conditioned neural network. To enhance sampling quality, they applied an \textit{exponential moving average} (EMA) of network parameters and carefully designed the step size schedules for Langevin dynamics, ensuring theoretical stability and empirical success in high-resolution image generation.

Jolicoeur-Martineau et al.~\cite{jolicoeur2020adversarial} further advanced this framework by incorporating an \textit{adversarial objective} into score-based models alongside traditional denoising score matching. Their approach introduces \textit{Consistent Annealed Sampling}, which stabilizes the sampling process and improves the quality of generated images. Experimental results demonstrate that their model produces samples with higher perceptual quality while preserving diversity. In another innovative direction, De Bortoli et al.~\cite{de2021diffusion} reinterpret score-based generative modeling as a solution to the \textit{Schrödinger Bridge Problem}, proposing a method based on \textit{Iterative Proportional Fitting (IPF)}. Their work connects probabilistic modeling with optimal transport theory, offering a principled framework for learning time-reversible diffusion paths between distributions.

Both DDPMs and SGMs can be unified under a continuous-time framework using \textit{stochastic differential equations (SDEs)}~\cite{yang2023diffusion}. This continuous generalization allows the diffusion and denoising processes to be described by SDEs rather than discrete Markov chains, enabling flexible noise scheduling and principled sampling. The general form of a forward SDE used in Score-based models is given by:
\begin{equation}
    dx = f(x, t)dt + g(t)dw
    \label{SDE}
\end{equation}
where \(f(x, t)\) is the drift coefficient, \(g(t)\) is the diffusion coefficient, and \(w\) denotes the standard Wiener process.

Specific forms of this SDE recover known diffusion models:
\begin{itemize}
    \item For DDPMs, the forward SDE takes the form:
    \begin{equation}
        dx = -\frac{1}{2}\beta(t)x dt + \sqrt{\beta(t)} dw
        \label{eq:SDE-DDPM}
    \end{equation}
    where \(\beta(t)\) denotes the time-dependent noise rate.
    \bigskip
    \item For SGMs, where the noise schedule is defined by \(\sigma(t)\), the SDE becomes:
    \begin{equation}
        dx = \sqrt{\frac{d[\sigma(t)^2]}{dt}} dw
    \end{equation}
\end{itemize}
To reverse the generative process, reverse-time SDEs and their deterministic counterparts, known as probability flow ODEs, are employed. These processes effectively transform pure noise into data samples while preserving marginal densities at each time step~\cite{song2020score}.

Despite their strengths, both DDPMs and SGMs face limitations, such as slow sampling speeds and sensitivity to noise schedules. These issues continue to motivate the development of more efficient and expressive variants, as discussed in the next section.

\subsection{Challenges and Limitations}

DMs exhibit remarkable generative capabilities in modeling complex data distributions, encompassing applications such as image synthesis, speech, and video. However, the formidable challenges of slow and resource-intensive training and sampling processes impede broader utilization of DMs~\cite{wu2023fast}. While existing solutions, including loss re-weighting and neural network refinement, are effective, they are post-hoc modifications that do not delve into the intrinsic mechanism of DMs. The key drawback of DMs persists in their requirement to execute multiple steps during inference to generate a single sample. Despite considerable research efforts towards this, GANs still outpace DMs in the speed of image production~\cite{croitoru2023diffusion}. To minimize uncertainty, DMs typically refrain from taking substantial steps during the sampling process. Employing smaller steps ensures that each step's generated data sample aligns with the learned Gaussian distribution. This pattern mirrors the behavior observed in the optimization of neural networks using gradient descent. Indeed, making a large negative step in the direction of the gradient, signifying a very high learning rate, may result in updating the model to a region characterized by high uncertainty, where the loss value becomes unpredictable and uncontrollable. In future research, incorporating update rules inspired by efficient optimizers into diffusion models may offer a pathway to enhance the efficiency of the sampling process~\cite{croitoru2023diffusion}.
\section{Hybrid Approaches}
\label{Hybrids}

Hybrid methodologies in generative modeling represent a sophisticated approach that integrates multiple distinct generative frameworks, such as Variational Autoencoders (VAEs) and Generative Adversarial Networks (GANs). The core motivation behind creating these hybrid architectures is to synergistically combine the advantageous characteristics of each constituent model while simultaneously mitigating their inherent limitations. VAEs, for instance, are typically recognized for their stable training dynamics and their ability to learn meaningful, structured latent representations of data, but often produce samples that lack the sharpness and fine detail seen in real data (i.e., they can be blurry). Conversely, GANs excel at generating high-fidelity, sharp samples that closely resemble real data distributions but are notoriously difficult to train, often suffering from issues like mode collapse (where the generator produces only a limited variety of outputs) and training instability.

By fusing these paradigms, hybrid models aim to achieve a more robust, versatile, and powerful generative framework. This section delves into significant developments and techniques in the construction of effective hybrid models, particularly focusing on the combination of VAEs and GANs, given their prominence and the complementary nature of their strengths and weaknesses. We will explore key architectures, the rationale behind their design, and the challenges they entail.
 \par


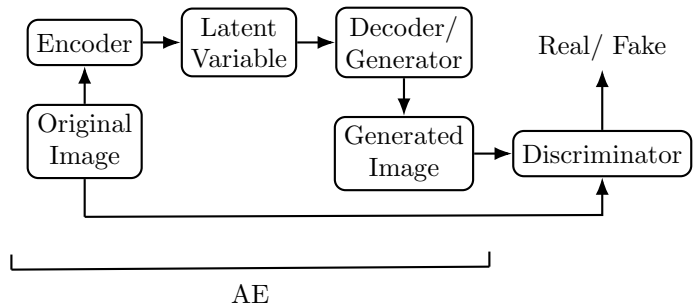
\begin{figure}[ht]
    \centering
    \begin{tikzpicture}[
    >=Latex, 
    node distance=0.5cm and 0.5cm, 
    block/.style={ 
        rectangle,
        draw,
        thick,
        text centered,
        rounded corners,
        minimum height=0.6cm,
        minimum width=1.5cm,
        align=center 
    },
    conn/.style={->, thick}, 
    span/.style={thick}     
  ]

  \node[block] (enc) {Encoder};
  \node[block, right=of enc] (lat) {Latent \\ Variable}; 
  \node[block, right=of lat] (decgen) {Decoder/ \\ Generator};

  \node[block, below=of enc] (orig) {Original \\ Image};
  \node[block, below=of decgen] (gen) {Generated \\ Image};
  \node[block, right=of gen] (disc) {Discriminator};
  \node[above=0.8cm of disc] (output) {Real/ Fake}; 

  \draw[conn] (orig) -- (enc);
  \draw[conn] (enc) -- (lat);
  \draw[conn] (lat) -- (decgen);
  \draw[conn] (decgen) -- (gen);
  \draw[conn] (gen) -- (disc);
  \draw[conn] (disc) -- (output);

  \coordinate[below =0.5cm and -0.2cm of orig] (orig_out); 
  \coordinate[below =0.5cm and -0.2cm of disc] (disc_in);  
  \draw[thick] (orig.south) -- (orig_out);
  \draw[conn] (orig_out) -- (disc_in) -- (disc.south); 

  \coordinate (ae_start) at ($(orig.south west) + (-0.2cm, -1.2cm)$);
  \coordinate (ae_end)   at ($(gen.south east) + (0.2cm, -1cm)$);
  \draw[span] (ae_start) -- (ae_end);
  \draw[span] (ae_start) -- ++(0, 0.2cm); 
  \draw[span] (ae_end)   -- ++(0, 0.2cm); 
  \node[below=0.1cm of {$(ae_start)!0.5!(ae_end)$}] {AE}; 

  \coordinate (gan_start) at ($(lat.south west) + (-0.2cm, -2.7cm)$); 
  \coordinate (gan_end)   at ($(output.east) + (0.5cm, -2.7cm)$); 

\end{tikzpicture}
    \caption{Overview of a hybrid network. Combination of VAE with GAN by collapsing decoder and generator into one.}
    \label{fig:VAE_GAN}
\end{figure}


\subsection{Hybrid GAN-VAE Architecture}

A foundational step in VAE-GAN hybridization was presented by Larsen et al. in 2016 \cite{larsen2016autoencoding}. Their innovative model integrated a GAN structure within the VAE framework by collapsing the decoder and generator into a unified VAE-GAN architecture, as illustrated in Figure~\ref{fig:VAE_GAN}. The key insight of their approach was to repurpose the learned feature representations from the GAN's discriminator to enhance the VAE's reconstruction process.

Standard VAEs typically measure reconstruction quality using element-wise distance metrics (like Mean Squared Error - MSE) between the original input ($x$) and the reconstructed output ($\tilde{x}$). While computationally simple, these metrics often fail to capture perceptual similarity effectively, contributing to the characteristic blurriness of VAE-generated samples.

The Larsen et al. model ingeniously replaces this element-wise reconstruction error. Instead, it utilizes the intermediate feature activations within the GAN's discriminator. The discriminator, trained to distinguish real from generated samples, inherently learns a rich hierarchy of features that are highly relevant to data realism and perceptual quality. By comparing the discriminator's feature representations of the original input ($\text{Dis}_l(x)$) and the reconstructed output ($\text{Dis}_l(\tilde{x})$) at a specific layer $l$, the model can employ a more perceptually meaningful similarity metric.

Specifically, they defined a Gaussian observation model for the discriminator's features at layer $l$:
\begin{equation}\label{eq:obs_model} 
p(\text{Dis}_l(x)|z) = \mathcal{N}(\text{Dis}_l(x) | \text{Dis}_l(\tilde{x}), I) 
\end{equation} 
Here, $z$ is the latent code from which the reconstruction $\tilde{x}$ is generated ($\tilde{x} = \text{Decoder}(z)$). This equation models the likelihood of observing the original image's features ($\text{Dis}_l(x)$) given the latent code $z$, assuming these features follow a Gaussian distribution centered around the reconstructed image's features ($\text{Dis}_l(\tilde{x})$) with an identity covariance matrix ($I$). A higher likelihood implies that the reconstructed image is perceptually similar to the original, as judged by the discriminator's learned features.

This likelihood is then incorporated into the VAE's objective function (specifically, the Evidence Lower Bound or ELBO) by replacing the standard reconstruction term with a negative log-likelihood term based on these discriminator features:
\begin{equation}\label{eq:dis_llike} 
L_{\text{Dis}_l \text{llike}} = -\mathbb{E}_{q(z|x)}[\log p(\text{Dis}_l(x)|z)] 
\end{equation} 
This term, $L_{\text{Dis}_l \text{llike}}$, encourages the VAE's encoder ($q(z|x)$) and decoder to produce reconstructions whose discriminator features closely match those of the original input.

The entire hybrid model is trained end-to-end using a combined loss function that balances three objectives:
\begin{equation}\label{eq:combined_loss} 
L = L_{\text{prior}} + L_{\text{Dis}_l \text{llike}} + L_{\text{GAN}} 
\end{equation} 
Where:
\begin{itemize}
    \item $L_{\text{prior}}$ is the standard VAE Kullback-Leibler (KL) divergence term ($D_{KL}(q(z|x) || p(z))$), which regularizes the latent space by encouraging the encoded distribution $q(z|x)$ to match the prior distribution $p(z)$ (often a standard Gaussian).
    \item $L_{\text{Dis}_l \text{llike}}$ is the novel discriminator-feature-based reconstruction loss defined above.
    \item $L_{\text{GAN}}$ is the standard GAN adversarial loss, applied to the VAE's decoder (acting as the generator) and the discriminator. This pushes the decoder to produce samples that the discriminator finds indistinguishable from real data.
\end{itemize}
By employing this learned similarity measure derived from the discriminator, this hybrid VAE-GAN architecture demonstrated the ability to generate image samples with significantly improved visual fidelity compared to traditional VAEs, effectively mitigating the blurriness issue.

\subsection{Adversarially Learned Inference (ALI) and Bidirectional GANs}

Concurrent with the VAE-GAN work, Dumoulin et al. introduced Adversarially Learned Inference (ALI) \cite{dumoulin2016adversarially}, also known as Bidirectional GAN (BiGAN). While not strictly a VAE-GAN in the sense of combining their loss functions directly, ALI/BiGAN introduced an adversarial framework that jointly learns a generative network (mapping latent codes $z$ to data $x$) and an inference network (mapping data $x$ to latent codes $z$).

The core idea involves a discriminator that learns to distinguish between pairs of (data sample, latent code). It is trained to classify jointly drawn pairs from the encoder distribution ($x \sim p_{\text{data}}$, $z \sim \text{Encoder}(x)$) as `real' and pairs from the generator distribution ($z \sim p_z$, $x \sim \text{Generator}(z)$) as `fake'. The generator and encoder are trained adversarially against this discriminator.

This adversarial process forces the learned inference mechanism (Encoder) to produce latent distributions that are indistinguishable from the prior distribution used by the Generator, and vice-versa. Key advantages highlighted by ALI include:
\begin{itemize}
    \item Improved Mode Coverage: Experiments showed ALI could capture complex data distributions with multiple modes more effectively than standard GANs, which sometimes collapse to generating only a few modes.
    
    \item Coherent Learning: The simultaneous, adversarial learning of generation and inference was found to be more effective than approaches where an inference network is trained separately after the generator is fixed.
    \item Competitive Performance: ALI demonstrated strong results on tasks like semi-supervised learning \cite{dumoulin2016adversarially}, indicating the utility of the learned inference network for downstream tasks.
\end{itemize}
ALI represents a significant step in equipping GANs with an explicit and efficiently learned inference mechanism, a feature naturally present in VAEs.

\subsection{Information-Theoretic Hybrids: InfoVAE}

Addressing the limitations of both VAEs (sample quality) and GANs (training stability, lack of inference) from a different angle, Zhao et al. proposed InfoVAE \cite{zhao2017infovae}. This model draws inspiration from both VAE principles and the information-theoretic concepts popularized by InfoGAN \cite{chen2016infogan}. InfoGAN modifies the standard GAN objective to maximize the mutual information between generated samples and a subset of the latent variables, encouraging the learning of disentangled and interpretable representations.

InfoVAE adapts this philosophy to the VAE framework. It aims to learn a robust generative model while simultaneously extracting informative latent features by explicitly maximizing the mutual information ($I(x; z)$) between the observed data ($x$) and the latent variables ($z$). The InfoVAE objective often involves modifications to the standard VAE ELBO, potentially incorporating alternative divergence measures like the Maximum Mean Discrepancy (MMD) or an adversarial loss inspired by GANs to better match the aggregated posterior distribution $q(z) = \mathbb{E}_{p(x)}[q(z|x)]$ to the prior $p(z)$, rather than solely relying on the instance-wise KL divergence in the standard ELBO.

By focusing on maximizing mutual information and employing architectures like DC-GAN for its network components, InfoVAE strives to:
\begin{itemize}
    \item Generate high-quality, consistent samples.
    \item Address VAE issues like latent variable variance overestimation ("posterior collapse," where the KL term dominates and the decoder ignores $z$).
    \item Learn more informative latent representations compared to standard VAEs.
    \item Bridge the gap between VAE and GAN methodologies through an information-theoretic lens.
\end{itemize}

\subsection{Self-Supervised Adversarial Hybrids: SS-AVL}

More recent hybrid approaches, such as Self-Supervised Adversarial Variational Learning (SS-AVL) by Ye et al. \cite{ye2023self}, integrate concepts from adversarial learning, VAEs, and self-supervised learning. SS-AVL proposes a framework where the generator and inference models are trained somewhat separately but synergistically. Key features of the SS-AVL approach include:
\begin{itemize}
    \item Factorized Latent Space: The generator, denoted $\mathcal{G}_\psi(z, d, c)$, utilizes a structured latent space comprising different types of variables: continuous ($z$), discrete ($d$), and potentially categorical or other continuous factors ($c$), sampled from independent priors. This allows for modeling different aspects of the data variation.
    \item Adversarial Generator Training: The generator is trained using adversarial learning to align its output distribution $P_\mathcal{G}$ with the real data distribution $P_\mathbf{x}$. It employs the Wasserstein GAN (WGAN) loss with a gradient penalty (WGAN-GP) term. This choice promotes training stability and helps prevent mode collapse compared to the original GAN formulation.
    \item Self-Supervised Inference Training: Crucially, the inference models (encoders mapping $x$ back to the latent factors $z, d, c$) are trained using a self-supervised learning objective derived from maximizing the log-likelihood, akin to the VAE framework. Lower bounds based on Jensen's inequality are used, potentially optimizing separate inference networks for the different types of latent variables (continuous and discrete/categorical). This self-supervised objective aims to learn meaningful representations without direct supervision.
    \item Separation of Objectives: By using adversarial loss primarily for the generator and a VAE-like/self-supervised loss for the inference networks, SS-AVL attempts to gain stability and leverage the representational power of self-supervision, potentially offering advantages over models where VAE and GAN losses are more directly combined or conflicted.
\end{itemize}

\subsection{Challenges and Limitations}
In representation learning, hybrid models such as GAN-VAE are prevalent. However, a persistent challenge in existing models arises from the joint training of the generator and inference modules, leading to issues with pre-trained GANs and accessing real data. One drawback of these generative models is the requirement for real samples in mutual information maximization, potentially resulting in the forgetting of past knowledge as mentioned by Ye \textit{et al.} \cite{ye2023self}. Additionally, ongoing issues such as unstable GAN training and the necessity to generate more diverse images still persist in this approach.
Another limitation is observed in hybrid approaches like Hierarchical Patch VAE-GAN \cite{gur2020hierarchical}. These models face a shortfall in semantic understanding, leading to potential difficulties in accurately representing the cohesive global structure of images in the output. This deficiency becomes evident, for instance, when local elements like people walking or cars moving are preserved, but the overall arrangement of scenes may appear unnatural.
We explore DMs next, a novel approach that help mitigate some of these issues such as unstable training.

\section{Application and Impacts}
\label{apps}
Generative AI and DL have profoundly influenced various domains, transforming task execution and creating novel possibilities for innovation and creativity. We explore diverse, transformative applications of generative AI in computer vision, content creation, healthcare, autonomous systems, education and training, data augmentation and synthesis, environmental modeling, simulation, and robotics and humanoid systems~\cite{chinta2024fairaied,chinta2025ai}.

\subsection{Data Augmentation and Synthesis}
\label{augmentation}
The importance of having an adequate volume of data for optimal neural network training cannot be overstated. Inadequate data leads to uncertain network parameters and poor generalization of learned networks \cite{alzubaidi2023survey}. Conventional methods for generating alternative data have limitations in promoting diversity~\cite{antoniou2017data}.
Generative AI, especially GANs, has revolutionized data augmentation by generating synthetic data samples, effectively addressing data inadequacy issues, enhancing diversity, and empowering models to learn from a more comprehensive and representative dataset, improving overall performance and generalization capabilities~\cite{Raner_Joshi_Sawant_2023}.
One of the GAN variants, called DAGAN,~\cite{antoniou2017data}, utilizes image-conditional GANs to generate additional data samples within the same class, extending its capabilities to encompass unseen classes. This approach significantly enhances classifier performance, even when combined with standard data augmentation techniques. DAGAN's versatility across various models and methods renders it highly valuable in low-data settings. As a result, DAGAN offers a powerful solution to the challenges posed by limited data availability, empowering neural networks to excel in challenging scenarios.

Diverse image styles are crucial in data augmentation. Methods like cropping, rotating, and flipping, combined with GANs capabilities~\cite{perez2017effectiveness}, effectively address this challenge.
Classification can be improved through a data augmentation technique known as neural augmentation~\cite{perez2017effectiveness}. By incorporating two augmentation methods, using GANs and basic transformations before training, and leveraging a pre-trained neural network for augmentation, the classification accuracy and overall performance are significantly enhanced. Particularly useful for tasks with limited data, these methods also hold promise in video data augmentation and addressing safety challenges in self-driving vehicles, as we shall see later.

\subsection{Autonomous Systems}
\label{Autonomous}

Self-driving cars and robots rely on realistic sensory inputs, particularly visual data, for navigation and interaction. Generative AI, crucial in self-driving vehicles, utilizes models to learn from extensive data, enabling autonomous responses to complex driving situations during training and simulation. The spatially-adaptive normalization (SPADE) technique, introduced by Park \textit{et al.}~\cite{park2019semantic}, incorporates semantic input layouts into affine transformations within normalization layers. This approach enhances the synthesis of realistic images across diverse scenes (\textit{e.g.}, indoor, outdoor, landscape, and street scenes). SPADE's versatility is evident in its effectiveness for multi-modal synthesis and guided image synthesis, making it applicable to various domains and creative tasks. The SPADE generator, a lightweight conditional image synthesis model, simplifies architecture by eliminating the encoder used in recent models. It employs ResNet blocks with upsampling layers and utilizes learned modulation parameters from SPADE for label layout information encoding, eliminating the need for segmentation maps in the first layer. This design enables multi-modal synthesis with a random vector input. The generator is trained using a multi-scale discriminator and a loss function based on pix2pixHD, where the least squared loss term is replaced by the hinge loss term. Semantic mask downsampling is employed to match spatial resolutions across scales.

An essential aspect of autonomous driving involves predicting future events and distinguishing various elements in its environment through self-supervision. This challenge is addressed by a generative model called GAIA-1 \cite{hu2023gaia}. GAIA-1 utilizes vector-quantized representations and can create realistic driving scenarios by incorporating video diffusion models, functioning as a sophisticated neural simulator. Its multimodal approach allows it to control the ego-vehicle's actions and scene attributes through textual and action-based instructions. Although promising, its autoregressive generation process currently operates slower than real-time, though parallelization can address this limitation. GAIA-1's significance is in its contribution to developing autonomous systems that can comprehend and adapt to real-world complexities, thereby improving their decision-making and generalization capabilities. Further, GAIA-1 is a valuable tool for generating diverse data for training and validating autonomous driving systems, including adversarial examples.

\subsection{Computer Vision}

Generative models have significantly influenced the field of computer vision (CV), with pivotal contributions from GANs, DMs, and VAEs playing crucial roles in addressing specific challenges, thereby fostering advancements in image generation, editing, and comprehension.
GANs, in particular, have been instrumental in image-to-image translation tasks, exemplified by their success in cartographic-to-aerial and monochromatic-to-color conversion. Variants like Conditional GANs (cGANs) \cite{isola2017image} have shown the ability to produce realistic images through L1 regression, surpassing conventional models. An enhanced variant of GANs, IDSGAN \cite{lin2022idsgan}, addresses convergence and instability concerns through the WGAN framework.
MTGAN \cite{zhang2020multi} addresses challenges related to false positives and small object detection when compared to CNNs, and has gained recognition for its exceptional image generation quality and improved detection accuracy. These GAN variants contribute to enhancing stability and proficiency, rendering GANs invaluable in object detection and tracking. 


Furthermore, VAEs have become strong competitors to GANs, benefiting from advancements that have improved their capabilities and addressed 
limitations in reproducing high-quality images. This has expanded the applications of VAEs beyond traditional generative tasks. The Deep Recurrent Attentive Writer (DRAW) model \cite{gregor2015draw} is noteworthy for its ability to generate intricate and realistic images, excelling particularly in image classification tasks. DRAW has similarities with VAEs, using recurrent networks for both the encoder and decoder. In contrast to other generative models, DMs, as illustrated by Croitoru et al. \cite{croitoru2023diffusion}, share key 
traits with VAEs, involving the mapping of data to a latent space and deriving an objective function as a lower bound of data likelihood. Further, compared to the commonly used GANs, DMs offer stable training and increased diversity in generated outputs, albeit at the cost of somewhat reduced efficiency during inference \cite{croitoru2023diffusion}. DMs have emerged as a promising option with enhanced performance in CV
especially in generating digital artworks and creative domains like the production of artistic paintings, text-guided image editing, video creation from text \cite{zhang2023text}, and even 3D object generation \cite{jabbar2021survey}. Notable examples include Multimodal Guided Artwork Diffusion (MGAD) \cite{huang2022draw} and DiffStyler \cite{huang2022diffstyler}, each integrating multimodal guidance and a controllable dual DM to preserve global content and boost creativity \cite{huang2022draw}. 

\subsection{Robotics and Humanoid Systems}
Recent robotics research focuses on creating personal assistant robots with human-like communication skills, including non-verbal elements like diverse gestures. However, developing a robot capable of expressing various human gestures remains a challenge \cite{arjovsky2017wasserstein}.
The modeling of human behavior during interactions involves proposing a method that integrates three crucial factors such as interaction intensity, time evolution, and time resolution into the network structure. This enables the capture of nuanced interaction motions using a deep generative model \cite{nishimura2020human}. The generative model is trained with WGAN-GP~\cite{arjovsky2017wasserstein}, a stable learning approach to help avoid mode collapse. Subjective evaluations indicate that the proposed method can generate high-quality human-like motions during interactions. Experimental results compare the proposed method to a baseline linear probabilistic principal component analysis (PPCA). It highlights the successful formulation of modeling interaction behavior through a deep generative GAN-based model, emphasizing the model's ability to generate high-quality interaction-like motions. Another key challenge is navigating the complexities of high-dimensional robotics systems.
The GAN framework has emerged as a solution to this, employing the ability to learn and generate valid configurations under constraints~\cite{lembono2021learning}.
Tailored for robotics, it incorporates extra inputs, costs, output augmentation, and an ensemble of networks. The approach speeds up inverse kinematics for high degrees of freedom (DoF) systems and accelerates constrained motion planning algorithms. GAN-generated samples enhance efficiency in numerical optimization for inverse kinematics and replace uniform sampling in motion planning, significantly reducing computation time. Validated in simulation with a 7-DoF Panda manipulator and  28-DoF humanoid robot Talos, this GAN-based 
approach is a 
promising 
solution to enhance the computational efficiency of robotic systems in complex environments~\cite{lembono2021learning}.

Robot training often requires human involvement to provide demonstrations, create simulation environments, and define reward functions for reinforcement learning (RL). This human involvement becomes a bottleneck, hindering the scalability of robot learning across various tasks and environments. To overcome these challenges, Gen2Sim~\cite{katara2023gen2sim}, a method that automates the generation of 3D assets, task descriptions, task decompositions, and reward functions using pre-trained generative models of language and vision is proposed. Gen2Sim employs image diffusion models to generate 3D assets from object-centric images and predicts physical parameters using LLMs.
The model directs LLMs to create task descriptions and rewards for assets. RL policies are then trained in generated environments using these rewards. Demonstrating applicability, a digital twin is built in simulation, and the trained trajectory is deployed in the real world. This streamlines robot skill learning in simulation, minimizing human intervention and automating training processes.There are also challenges in robust manipulation in robots when integrating DL and RL~\cite{liu2021deep}. However, drawing inspiration from methods such as imitation learning, GANs, and meta-learning can help address issues related to data dimensionality and scalability. This approach is suited for tasks with sparse reward signals, like robotic manipulation and control, improving sample efficiency and generalization.


\subsection{Healthcare}
Recent breakthroughs in generative AI have revolutionized the field of medical imaging, leading to a range of applications such as anomaly detection~\cite{pinaya2023generative}, image-to-image translation employing CycleGAN \cite{zhu2017unpaired}\cite{pinaya2023generative}, de-noising, and magnetic resonance imaging (MRI) reconstruction~\cite{pinaya2023generative}.
A framework called MONAI generative models \cite{pinaya2023generative}, an open-source platform for researchers and developers, was introduced to facilitate the straightforward training, evaluation, and deployment of generative models and their associated applications. 
The platform replicates state-of-the-art studies in a standardized manner, incorporating various architectures like DMs, autoregressive transformers, and GANs.
Adaptive Diffusion Priors (AdaDiff)~\cite{gungor2023adaptive}, an innovative technique for MRI reconstruction, unlike traditional methods with static diffusion priors, adjusts its priors during inference to better align with test data distribution, enhancing the authenticity of generated images. Furthermore, the task of predicting protein structures from a given fixed protein sequence is made possible through the introduction of a novel diffusion model~\cite{jing2023eigenfold}. Different generative models have significantly contributed to many aspects of healthcare, spanning from clinical record-keeping ~\cite{shokrollahi2023comprehensive}, aiding in diagnoses, interpreting radiology results, supporting clinical decisions, managing medical coding and billing \cite{shokrollahi2023comprehensive}, and drug development and molecular representation~\cite{shokrollahi2023comprehensive}.\par

Despite these advancements, the ongoing pursuit of domain generalization, model robustness, and fairness in medical imaging underscores the commitment to addressing challenges and ensuring the ethical deployment of artificial intelligence in healthcare. Notably, the autonomous generation of realistic data augmentations by generative models has emerged as a powerful strategy, driving improvements in prediction accuracy and fairness across diverse medical modalities. As we navigate the evolving landscape of healthcare applications, the integration of generative models stands as a testament to the boundless possibilities they offer in enhancing both efficiency and equity within the realm of medical practice~\cite{ktena2023generative}.

\subsection{Environmental Modeling and Simulation}

Simulation models serve as experiments in understanding past ecosystems where direct manipulation is impossible. Palaeoecology and archaeology face challenges beyond data collection, requiring innovative synthesis methods. Generative models, like agent-based models (ABMs), go beyond traditional simulations, offering a bottom-up approach with explicit representation of human decision-making. ABMs, although underused in ecological reconstructions, provide a flexible framework capturing reciprocal human-environment feedback. Despite the potential of generative models, they're often underutilized in palaeoecology~\cite{perry2016experimental}. The prevalent focus on model validation using historical data has limitations, hindering predictions of future scenarios. An integrated, bidirectional approach, where models inform data interpretation and vice versa, holds promise for deeper insights into both past ecosystems and modeling techniques.
Moreover, environmental modeling and simulation benefit greatly from remote sensing image understanding. 
Valuable insights about the Earth’s surface can be extracted by analyzing remote sensing data, which allows understanding and predicting environmental behavior in various scenarios~\cite{liang2020deep}.\par

A novel approach is proposed to address the challenges in real-time fire nowcasting, with specific focus on the recent surge in wildfires worldwide. Utilizing a 3D Vector-Quantized VAE (3DVQ-VAE) \cite{cheng2023generative}, the model surpasses computationally intensive physics-driven models like cellular automata and computational fluid dynamics. Unlike ML models requiring region-specific data, VQ-VAE discretizes the latent space, yielding more coherent wildfire spread sequences. With a trainable codebook and three key loss functions, VQ-VAE efficiently generates new fire spread scenarios in a given ecoregion, leading to a leap in wildfire prediction efficiency.

\subsection{Content Creation}
In recent years, GANs have significantly advanced content creation techniques. One notable contribution is the Dynamic ResBlocks GAN (DRB-GAN) introduced by Xu \textit{et al.}~\cite{xu2021drb} to specifically address artistic style transfer through three interconnected networks: style encoding, style transfer, and a discriminative network. By incorporating \textit{style codes} as shared parameters and utilizing Dynamic ResBlocks, DRB-GAN effectively bridges the gap between arbitrary and collection style transfer. The model has demonstrated efficiency in high-resolution transfers, resilience across resolutions, and effectiveness with unseen styles, as evidenced by qualitative comparisons with superior performance over baseline models.
An art-generating agent, aiming to enhance the arousal potential in the generated art has been developed by Eglemma \textit{et al.} \cite{elgammal2017can}.
This modified GAN operationalizes creativity by maximizing deviation from established styles while minimizing deviation from the art distribution. The agent's objective is to produce art that is both novel and within a specified range, striking a delicate balance between novelty and familiarity.

Another notable model is GPT-4~\cite{openai2023gpt},  which utilizes GANs to create novel textual data, is a cutting-edge multimodal model proficient in processing both image and text inputs to generate text outputs. It has achieved human-level performance on professional and academic benchmarks, including a top 10\% score in a simulated bar exam~\cite{openai2023gpt}. Leveraging a Transformer-based architecture, GPT-4 undergoes pre-training and post-training alignment processes to enhance factual accuracy and behavior adherence. Furthermore, VAEs find application in establishing a precise connection between natural language and the visual content of images, leading to a comprehensive understanding of images. A symmetric joint embedding model that facilitates image retrieval based on textual queries and enhances zero-shot accuracy is constructed using VAEs \cite{reed2016learning}. Additionally, the complex task of Question Answering in natural language processing is tackled by introducing a dynamic memory network (DMN) architecture \cite{kumar2016ask}. The DMN processes questions and generates relevant answers by incorporating text perception and making inferences using VAEs based on pertinent facts. 

\section{Ethical Consideration}
\label{ethics}
The ethical implications of general AI have been extensively discussed, but ethical considerations specific to generative AI are a relatively new area of inquiry. 
This section critically examines the principal ethical challenges associated with generative AI, emphasizing its potential and the risks of misuse.

\subsection{Intellectual Property and Copyright}

The issue of intellectual property (IP) rights and copyright in generative AI involves uncertainties about ownership and usage rights for content produced by AI models. Determining rightful claimants and permissible utilization, especially for economic purposes, is a key challenge. The debate on copyright protection for AI-generated works centers on their eligibility for such protection~\cite{landes2003economic,xu2024transforming,yin2025digital}. The integration of machine-generated artworks into the contemporary art scene highlights a series of legal intricacies, with a pivotal inquiry centering on the permissibility of using copyrighted materials as training sets for generative models~\cite{franceschelli2022copyright}. As these models gain prominence, questions abound regarding the legal protocols for storing copies during the training process and, notably, the rightful ownership of copyrights pertaining to the generated output. This narrative extends seamlessly into the realm of code generation, with GitHub Copilot triggering significant debate surrounding copyright implications~\cite{franceschelli2022copyright}. Copilot's capacity to autonomously generate code from publicly available sources introduces myriad legal complexities, prompting a nuanced exploration from the framework of IP law, encompassing considerations of fair use doctrine, potential modifications to licensing agreements, and ethical dimensions, offering valuable insights for ML researchers at the intersection of artistic innovation, technological progress, and copyright regulation.

Another challenge is implementing appropriate safeguards to respect the rights of original creators whose creations generative AI may have been trained on while allowing for new creations. To achieve this, it is key to train models that learn abstract and non-copyrightable features from training data, rather than simply memorizing a condensed version of it. A key concern with generative AI is whether everyday uses of the technology may produce outputs that infringe on specific copyright interests. Matthew Sag proposed an approach to determine the likelihood of infringement based on factors such as the probability of memorization, the number of duplicates of a work, and ratio of model size to training data~\cite{sag2023copyright}. 

The absence of uniform laws exacerbates concerns about unauthorized AI model use and distribution~\cite{poland2023generative}. The case of \textit{Kashtanova and Zarya of the Dawn} exemplifies a copyright challenge stemming from the use of generative models~\cite{poland2023generative}. In a landmark decision on February 21, 2023, the US Copyright Office recognized that the collaborative work involved co-authorship between Kristina Kashtanova and Midjourney's AI technology. Despite the AI's role in generating images for the work, the Copyright Office acknowledged that Ms. Kashtanova's contributions, encompassing the text, selection, coordination, and arrangement of written and visual elements, qualified for copyright protection. Notably, to secure this protection, Ms. Kashtanova had to disclaim any rights over the AI-generated content. This case underscores the evolving complexities in determining copyright ownership and protection when using generative models in creative endeavors~\cite{poland2023generative}.

\subsection{Bias and Fairness} 
\label{sec:fairness}

Addressing bias and boosting fairness in AI, especially in image and video analysis, is a formidable challenge. AI systems, especially those dealing with visual data, may pick up and perpetuate biases from historical or skewed training data. This issue is widespread, affecting facial recognition and video-based decision-making tools, leading to inaccuracies based on appearance or behavior~\cite{sixta2020fairface,gichoya2023ai,sham2023ethical}. A key line of research is the exploration of bias in myriad domains and different applications due to the use of AI. For example, in the domain of cardiac MR image segmentation, a study examining racial bias due to imbalanced training data found substantial racial biases in Dice performance metrics when using the nnU-Net~\cite{isensee2021nnunet} model on a diverse UK Biobank dataset~\cite{sudlow2015ukbiobank}. Interventions included stratified batch sampling, fair meta-learning with a race classifier, and developing segmentation models for each racial group. Results showed significant improvements in fairness, especially for protected racial groups. The study, focused on ED and ES short-axis cine cardiac MR images, showed pronounced racial bias with no significant gender bias detected.
Feng \textit{et al.}~\cite{feng2022has} investigated gender bias in image search engines such as Google, Baidu, Naver, and Yandex, along with proposed adversarial attack queries to assess the adequacy of gender bias mitigation. They found these attacks could effectively trigger high levels of gender bias. To mitigate this, they propose three re-ranking algorithms: the epsilon-greedy algorithm, the relevance-aware swapping algorithm, and the fairness-greedy algorithm. Experiments on simulated and real-world datasets showed that these algorithms effectively mitigate gender bias. 

Another creative solution introduced to reduce bias in AI is ``sketching'' in image classification~\cite{yao2022improving}, which transforms input images into sketches, preserving semantic information while eliminating bias. The model incorporates a fairness loss function, utilizing mean square error (MSE) loss of Statistical Parity Difference (SPD) along with classification loss for parameter optimization. Fairness is assessed using metrics like SPD, Equal Opportunity Difference (EOD), Difference in Equalized Odds (DEO), and Average Odds Difference (AOD). Validation on the CelebA dataset ~\cite{liu2015faceattributes} for face recognition and the Skin ISIC 2018 dataset~\cite{codella2018skin} for skin cancer classification demonstrates improved fairness and high accuracy. Furthermore, advancements such as fairness-aware facial image-to-image translation~\cite{hwang2020fairfacegan} signify progress in unbiased generative AI. In multimodal ML, especially in automated video interviews for employment assessment~\cite{booth2021bias,kim2023fairness}, ensuring fairness is crucial. An AI-based employment screening addressing this with a dataset of 4,255 videos from 733 undergraduates, employing three Random Forest models, found trade-offs between accuracy and bias reduction, stressing the challenges in achieving fairness in high-stakes AI uses like job interviews.

In addition, there are methods that focus on special image data, \textit{i.e.,} graph data. Current methods in graph structure generation mainly concentrate on predicting connections for generated nodes or reconstructing the input graph. However, these methods often struggle with an uneven distribution of node edges. For instance, NetGan~\cite{bojchevski2018netgan}, which performs random walks on the input graph, tends to reinforce biases present in the original data. Since the input graph often has more tightly linked nodes with the same sensitive attributes, NetGan learns and amplifies this bias, exacerbating group segregation in the generated graph. Similarly, TagGen~\cite{zhou2020data} faces a challenge where transfer probabilities of nodes can vary significantly across sensitive groups, leading to overrepresentation of highly connected nodes. To address these issues, $\rm{FG^2AN}$~\cite{wang2023fairness} aims to mitigate graph structure bias and connectivity bias. This ensures that the generated graph accurately represents each group without introducing additional biases. $\rm{FG^2AN}$ does this by carefully balancing the representation of different groups and controlling the distribution of edges, thereby fostering a more equitable representation in the generated graph structure. On the other side, FDGen~\cite{wang2025fdgen} extends the graph diffusion-based framework for fair graph generation, introducing a fairness regularizer that guides nodes to integrate representations from neighbors with different sensitive attributes while reducing the influence of neighbors with the same attribute. This approach addresses biases in the generated graphs while preserving the utility of the learned representations. Beside, the FG-SMOTE~\cite{wang2025fg} employs a synthetic minority oversampling strategy for fair node classification by generating new node samples and edges. It manages to keep node embeddings free of sensitive attribute effects, thus balancing node distributions and minimizing structural bias in the generated graph. Finally, $\rm{FS^2}$~\cite{wang2023preventing} introduces a fairness-aware generation model tailored for the streaming setting, which generates new decision instances sequentially while enforcing fairness constraints, thereby ensuring that both utility and fairness are maintained throughout the data stream.


%
\subsection{Deepfakes and Misuse}
The surge in generative AI, exemplified by deepfakes~\cite{george2023deepfakes}, has sparked notable ethical and privacy concerns. Deepfakes employ DL techniques to produce realistic synthetic media, manipulating images and videos by convincingly altering content or superimposing faces from one person onto another's body.
Modern deepfakes, relying on GANs \cite{goodfellow2014generative, george2023deepfakes}, involve the generator learning facial features from source media for facial deepfakes, while the discriminator identifies unnatural elements. With abundant training data, GANs convincingly transpose faces onto target footage, often utilizing techniques like facial motion modeling and voice cloning.
Apps like FakeApp and DeepFaceLab democratize deepfake creation, offering user-friendly interfaces and pre-trained models~\cite{Marr_2023}.\par
The challenge in verifying the authenticity of generated content and the potential for misuse make deepfake technology dangerous. Instances of malicious actors using deepfakes for harassment or economic gain have already emerged. Beyond IP and copyright issues, deepfakes possess the ability to inflict significant and challenging-to-address harm on individuals in society \cite{CliffordChance2020}.
For example, 
the digital recreation of Luke Skywalker in The Mandalorian's season 2 finale. Despite fans' initial excitement, the deepfake created by YouTuber Shamook raised concerns about the authenticity of the youthful Mark Hamill reconstruction~\cite{Baxter_2021}.
These models can spread misinformation, manipulate public opinion, and harm individuals through harassment or defamation. The slowed-down video of Nancy Pelosi demonstrated the potential for geopolitical manipulation, as the altered footage distorted her speech, highlighting the technology's misuse capacity~\cite{Goggin}.
Further, a video with Obama's impersonation using deepfakes showcased the risk of convincingly replicating public figures' speeches and mannerisms \cite{Choi_2022}. The viral video of Yang Mi placed in a 1983 Hong Kong TV series highlighted potential cultural and ethical implications, leading to its removal by Chinese authorities~\cite{ChiradeepBasuMallick2022}. Moreover, a Zuckerberg deepfake \cite{Goggin}, bragging about Facebook's control, underscored the deceptive possibilities of the technology, emphasizing the need for vigilance in discerning manipulated content. These instances illustrate the dual nature of deepfakes as an entertainment tool and a threat to truth and authenticity in various domains.

Spurred by misuse, researchers have been investigating methods to improve the detection of manipulated images of detecting the realism of advanced image manipulations. One notable study utilized an automated benchmark to detect facial manipulation, encompassing techniques such as DeepFakes, Face2Face, FaceSwap, and NeuralTextures, exploring various compression levels and sizes~\cite{rossler2019faceforensics++}. It demonstrated that current facial image manipulation methods can be detected effectively by trained forgery detectors, despite their visually impressive outcomes \cite{rossler2019faceforensics++}.
Learning-based approaches excel in challenging scenarios like low-quality videos, outperforming human observers and hand-crafted features in identifying manipulations. The study introduces a large dataset of manipulated face videos for forgery detection. Incorporating domain-specific knowledge improves detection, even under strong compression, surpassing human observers' capabilities. While current research emphasizes the importance of learning-based techniques and domain-specific knowledge to image manipulations successfully, ongoing research is crucial to stay ahead in the battle against sophisticated forgeries with the continued advancement of image-manipulation methods.\par


\section{Challenges and Future Directions}
\label{future}

Generative AI has made strides but faces challenges in training stability, mode collapse, scalability, and control. Efforts are needed to develop sophisticated image generation techniques and implement ethical safeguards. By delving into these challenges researchers can augment their understanding, leading to a future in which generative AI produces even more visually captivating and contextually-relevant content. Our exploration of challenges and potential future directions contributes to the advancement of generative models, providing researchers with an array of research problems to choose from.

\subsection{Improved Training Stability and Mode Coverage}
Training generative models especially GANs has challenges like mode collapse, where diverse samples are hard to generate, causing the exclusion of specific modes during training \cite{saxena2020generative}. Non-convergence and instability may happen, especially when one player dominates, leading to issues like vanishing gradients. The missing mode problem complicates GAN training as generated samples may not represent the entire data distribution well.
%
Mitigating the challenges inherent in GAN training necessitates more research endeavors, yielding re-engineered network architectures like cGANs~\cite{isola2017image}, LAPGAN~\cite{denton2015deep}, DCGAN ~\cite{radford2015unsupervised}, GAN-VAE~\cite{larsen2016autoencoding}, and InfoGAN~\cite{chen2016infogan}. Additionally, the incorporation of novel loss functions, exemplified by WGAN~\cite{arjovsky2017wasserstein}, LSGAN~\cite{mao2017least}, WGAN-GP~\cite{gulrajani2017improved}, and SN GAN~\cite{miyato2018spectral}, plays a pivotal role in addressing these challenges. This fusion of advanced architectures and refined loss functions contributes to the achievement of mode convergence and fosters stable training dynamics in GANs~\cite{saxena2020generative}.

Despite issues like unbalanced learning and overfitting, the DL community is actively looking for solutions. Researchers have tweaked GAN architectures in the past, and recent work is concentrating on regularization for better stability. The discriminator's sensitivity to data distribution and vulnerability to adversarial attacks emphasize the necessity for strong defense methods. Although there has been exploration into adversarial training, there isn't a foolproof defense, and training stability of GANs is not robust against various attacks. Looking ahead, researchers aim to improve GANs robustness and investigate optimization algorithms inspired by nature to tackle challenges in GANs optimization \cite{sajeeda2022exploring}.

\subsection{Enhanced Control and Interpretability}

Enhanced control and interpretability in Generative AI involve maintaining control over generated outputs and understanding AI models. While humans effortlessly connect data from different domains, AI needs numerous ground-truth examples to automatically uncover and interpret these connections. To overcome the need for expensive pairing, Kim \textit{et al.}~\cite{kim2017learning} tackled the challenge of discovering cross-domain relations using unpaired data. They propose a precise definition of cross-domain relations and introduce the challenge of learning to identify these relations between two distinct domains to propose a new method leveraging GANs called DiscoGAN~\cite{kim2017learning}. This network can transfer style from one domain to another effectively while retaining crucial attributes like orientation and facial identity, and the results have demonstrated that DiscoGAN can produce impressive images with style transfer capabilities. \par

Moreover, an analytic framework \cite{bau2018gan} is presented to visualize and comprehend GANs at unit, object, and scene-level to identify the interpretable structure and give insights into the internal mechanisms of a GAN. Interpretable units related to object concepts are identified, and the fundamental effect of these units is quantified through interventions. Through an examination of representation units, it is discovered that various aspects of GAN representations can be interpreted as signals correlated with object concepts and as variables influencing the synthesis of objects in the output. These interpretable effects offer valuable insights for comparing, debugging, modifying, and reasoning about GAN models. It is possible to apply this method to other generative models like VAEs and RealNVP~\cite{bau2018gan}.\par



\subsection{Real-Time and Interactive Generation}

Generative agents can exhibit real-time and interactive generation. Real-time responses in online forums create instant conversation threads, resembling discussions by stateless personas. Interactive generation adapts to users' environments, such as a generative agent mirroring Sal's daily life, a character from a popular vignette \cite{park2023generative}. This agent dynamically interacts with Sal's routines and technology, offering personalized experiences like brewing coffee and adjusting ambiance based on mood. Generative agents enhance technology experiences by adapting in real-time to users' behaviors and preferences. As an example of the practical benefits of real-time generative AI for effective interaction between users and AI system, we can discuss a live system generating gestures in response to speech \cite{rebol2021real} that uses GANs to understand the link between speech and gestures. Trained on a substantial volume of online speaker video data, it creates speaker-specific gestures from two-second audio segments. The system portrays these anticipated gestures on a virtual avatar, ensuring a seamless experience with less than a three-second delay.\par 
Simulating human behavior within interactive generation poses several challenges. The resource-intensive nature of simulations necessitates optimization for cost-effectiveness, indicating that exploring parallelization and specialized language models has been identified as a limitation in interactive simulations of human behavior. Recommendations for improved performance involve enhancing the retrieval module, particularly through the fine-tuning of functions. Evaluation limitations, such as a short timescale and dependence on a baseline human condition, call for extended observations and more rigorous benchmarks. Concerns regarding robustness, biases, and stereotypes in generative agents underscore the importance of comprehensive testing and ethical considerations in future research~\cite{park2023generative}.
\subsection{Generalization to Unseen Data}
In generative AI, the ability of a model to produce meaningful and coherent outputs for new, unseen data is known as generalization. VAEs achieve better generalization through techniques such as representation learning, regularization, data augmentation, and transfer learning. Using variational inference to learn a probabilistic latent space allows VAEs to generate diverse and realistic outputs beyond training data. On the other hand, adversarial training with GANs and careful monitoring enhance their capabilities in real-world applications.
One such research focuses on developing accurate and scalable probabilistic models for handling complex data distributions in generative models~\cite{rezende2014stochastic}, and introduce a class of deep directed generative models with Gaussian latent variables. Stochastic backpropagation is used to adjust model parameters, resulting in realistic samples and predictive capabilities for missing data.\par

The challenge of classifiers failing when encountering test data with different distributions from the training data can significantly impact detection tasks. To address this, the \textit{softmax prediction probability baseline} method \cite{hendrycks2016baseline}  aims to identify cases where classifiers make mistakes or encounter unseen examples, which is crucial for AI safety and reliable applications.
Additionally, the \textit{abnormality module} technique \cite{hendrycks2016baseline} is proposed, which outperforms the baseline in certain situations. This indicates the potential for further improvement in detecting errors and unusual inputs in ML-systems. Advancing these methods will enhance the reliability and effectiveness of machine learning classifiers in real-world scenarios.

\subsection{Ethical guardrails}
Delving into ethical considerations arising from generative AI, we focus on key areas such as intellectual property (IP) and copyright, as well as issues related to bias, fairness, and the potential for deepfakes and misuse. Examining the technical challenges associated with these considerations provides a comprehensive understanding of the complexities involved.

Globally, endeavors are underway to tackle the risks and concerns linked to the unlawful use of generative AI. In the European Union, discussions around the AI Act~\cite{hacker2023ai} include considerations for disclosure requirements regarding generative AI use and bans on applications such as biometric identification~\cite{bacalu2022biometric, milossi2021remote}, social scoring, and specific facial recognition and predictive policing systems. Meanwhile, in the United States, President Biden's Executive Order focuses on establishing standards and guidelines for the safe utilization of AI, including generative AI~\cite{TheWhiteHouse_2023}, with emphasis on key aspects such as safety, security, innovation, equity, consumer protection, privacy, federal government use, and global leadership.
Additionally, U.S. lawmakers are taking legal steps against issues like AI-generated pornographic content. The ``Preventing Deepfakes of Intimate Images Act'' \cite{HR9631} aims to prohibit the nonconsensual disclosure of digitally manipulated explicit content, with potential civil actions, fines, and imprisonment for offenders. Similar legislation has been enacted in New York, criminalizing the unlawful dissemination of intimate images created through digitization~\cite{Hinchey_2023}.
Working against the spread of disinformation through fake media, organizations are endorsing standards like the Coalition for Content Provenance and Authenticity (C2PA)~\cite{C2PA} and utilizing tools such as public key cryptography.
Notably, Leica's M11-P camera \cite{LeicaCamera} features content credentials for authentication. 

\subsection{Large Language Models in Image and Video Generation}

Large Language Models (LLMs) have shown promise for tasks like text-based image creation, but applying them to complex image or video generation presents open directions for further exploration~\cite{wang2024history}. The main challenge lies in aligning natural language instructions with visual outputs, which demands advanced multi-modal learning methods that can be computationally heavy~\cite{tang2025video}. Another area is the need to reduce the risk of biased or incorrect content, because LLMs may inadvertently inherit biases from their training corpora~\cite{chu2024fairness}. A third challenge is achieving fine-grained control over visual outcomes, given that LLMs typically parse text tokens without a built-in understanding of deeper visual structures~\cite{li2025benchmark}. Although some approaches combine LLMs with specialized decoders or cross-modal modules, ensuring high-quality and diverse outputs remains an open question. Future work could concentrate on robust training strategies, refined alignment techniques, and careful data handling to generate high-quality visuals while limiting undesired biases.

\section{Conclusion}
\label{conclusion}

The rapidly-evolving landscape of generative AI poses challenges in staying abreast of pertinent literature, hindering the identification of compelling avenues for future research. This paper undertakes a comprehensive survey of generative AI, with a particular focus on generative adversarial networks, variational autoencoders, their hybrids, and diffusion models, addressing their core limitations and stable training dynamics. Employing an innovative taxonomy, we systematically organize and summarize the literature, providing a critical analysis of relevant methodologies, their variations and enhancements, and persisting challenges. Furthermore, we delineate a diverse array of application domains for these methods, while also addressing the ethical considerations they entail. Finally, we discuss limitations and propose future directions, serving as valuable insights for future research endeavors.

\nocite{wang2023preventing,zhang2023individual,wang2023fg2an,wang2023mitigating,chinta2023optimization,chu2024fairness,yin2024improving,wang2023towards,wang2024toward,chinta2024fairaied,wang2024individual1,doan2024fairness1,wang2024advancing,wang2024group,yin2024accessible,wang2025fg,wang2025graph,wang2025fair,wang2025towards,yin2025digital,chinta2025ai,wang2025fdgen,wang2025Fairness,wang2025Redefining,zhang2019faht,zhang2024ai,zhang2022longitudinal,zhang2023censored,zhang2025fairness,zhang2022fairness,wang2024towards,saxena2023missed,zhang2019fairness,zhang2020flexible,zhang2020learning,zhang2021farf,zhang2021fair,zhang2023fairness,zhang2016using,zhang2018content,zhang2021autoencoder,zhang2018deterministic,tang2021interpretable,zhang2021disentangled,liu2021research,liu2023segdroid,cai2023exploring,guyet2022incremental,zhang2024fairness,wang2025FairnessT,zhang2025online,yinAMCR2025,Wang2025Unified,ijcai2025p64,ijcai2025p63,zhang2025datasets,WangAIFairness2025,palikhe2025towards,yin2025Uncertain}

\bibliography{sn-bibliography}

\end{document}